\def\archiveversion{}

\documentclass{article}
\usepackage{iclr2027_conference,times}
\ifdefined\archiveversion\iclrfinalcopy\fi

\usepackage{amsmath,amsfonts,amsthm,bm}

\def\eqref#1{equation~\ref{#1}}

\def\1{\bm{1}}

\DeclareMathAlphabet{\mathsfit}{\encodingdefault}{\sfdefault}{m}{sl}
\SetMathAlphabet{\mathsfit}{bold}{\encodingdefault}{\sfdefault}{bx}{n}

\usepackage{microtype}
\usepackage{graphicx}
\usepackage{subcaption}
\usepackage{booktabs}
\usepackage{longtable}
\usepackage{array}
\usepackage{enumitem}
\usepackage{xcolor}
\usepackage{algorithm}
\usepackage{algorithmic}
\usepackage[hidelinks]{hyperref}
\usepackage[capitalize,noabbrev]{cleveref}
\usepackage{amsmath}
\usepackage{amssymb}
\usepackage{mathtools}
\usepackage{amsthm}
\usepackage[textsize=tiny,disable]{todonotes}
\usepackage{placeins}
\usepackage{float}
\usepackage{wrapfig}

\theoremstyle{plain}
\newtheorem{theorem}{Theorem}[section]
\newtheorem{proposition}[theorem]{Proposition}

\theoremstyle{definition}

\theoremstyle{remark}

\crefname{assumption}{assumption}{assumptions}
\Crefname{assumption}{Assumption}{Assumptions}

\title{Compositional Self-Improvement: Promises and Pitfalls}
\ifdefined\archiveversion
\author{Changho Shin$^{1}$ \quad Daiwei Chen$^{2}$ \quad Brenden Lake$^{1,3}$ \\
\bf Frederic Sala$^{4}$ \quad Ramya Korlakai Vinayak$^{2}$ \\[4pt]
\footnotesize\normalfont $^{1}$Department of Computer Science, Princeton University, Princeton, NJ \\
\footnotesize\normalfont $^{2}$Department of Electrical and Computer Engineering, University of Wisconsin--Madison, Madison, WI \\
\footnotesize\normalfont $^{3}$Department of Psychology, Princeton University, Princeton, NJ \\
\footnotesize\normalfont $^{4}$Department of Computer Sciences, University of Wisconsin--Madison, Madison, WI}
\else
\author{Anonymous Authors}
\fi

\begin{document}

\maketitle
\ifdefined\archiveversion\lhead{}\fi

\begin{abstract}
Although self-improvement is becoming increasingly important as tasks become more complex, models struggle to generate reliable supervision for harder, out-of-distribution instances. One way to address this challenge is to exploit compositional structure: models often solve simpler subproblems more reliably than full problems, so their subproblem predictions can be composed into supervision for harder ones. We systematically study when this kind of composed supervision supports self-improvement and when it fails. Our controlled experiments identify three failure modes: unreliable subproblem predictions, errors introduced by composition, and limited supervision coverage. Beyond controlled settings, these findings carry over to more realistic coding and reasoning tasks: composed labels are more accurate than direct self-generated labels and improve performance even on compositions longer than those seen during training, while the failure modes remain visible. Further, although these failure modes often keep compositional self-improvement from fully succeeding on its own, a short SFT warmup with compositional self-improvement can bootstrap reinforcement learning on tasks that require composing skills the model already has, particularly when direct exploration rarely succeeds. Together, these findings characterize when compositional self-improvement can provide sufficiently reliable supervision to expand model capabilities.
\end{abstract}

\section{Introduction}

As models tackle increasingly complex tasks, reliable human supervision can become costly or difficult to obtain. This motivates methods that can improve models with less reliance on human annotations. One approach to \emph{self-improvement} is to learn from self-generated supervision, such as pseudo-labels or model-generated feedback \citep{lee2013pseudo,huang_2023_llms_self_improve,lee_2025_self_improving}. However, extending self-improvement beyond the range of problems the model can already solve reliably requires supervision on harder, out-of-distribution instances --- precisely where the model’s own predictions may be unreliable. 

We explore whether compositional structure can bypass this bottleneck. Even when a model cannot solve a complex problem directly, it may still be able to solve its components. This suggests a route to supervision: predictions on these components can be combined into a training target for the full problem. We study this iterative setting as \emph{compositional self-improvement}: the model constructs supervision for harder instances from predictions on simpler components, trains on the resulting targets, and repeats the process as its capabilities expand.

This approach builds on a broader literature using compositional structure to construct training data by recombining existing examples or their constituent parts, usually with the goal of improving generalization to novel compositions \citep{jia_liang_2016_recombination,andreas_2020_geca, qiu_2022_compositional_augmentation}. More recent work similarly composes existing problems into harder or longer-horizon instances for reasoning evaluation and training \citep{xiao_2025_a_plus_b,motwani_2025_h1,lu_2026_rhorizon, xu_2026_composition_rl}. Iterated amplification provides a closer conceptual precedent, constructing supervision for difficult problems from solutions to easier subproblems \citep{christiano_2018_amplification}.

Yet prior work leaves open when compositional supervision can sustain genuine self-improvement. Once the component solutions themselves are model-generated, errors can propagate through composition. Moreover, the composition rule can fail even with correct components, while filtering such errors can leave important regions without supervision.

We systematically investigate these challenges and identify three bottlenecks to continued compositional self-improvement: \emph{unreliable component predictions}, \emph{structural errors in composition}, and \emph{limited supervision coverage}. Each bottleneck calls for different remedies. Improving the underlying model reduces errors inherited from the components; repairing or restricting the composition rule addresses errors introduced during composition; and when filtering creates coverage gaps, further progress requires supervision that reaches the missing patterns. Thus, making composed labels more accurate is not by itself sufficient for continued self-improvement.

%
Our experiments illustrate both these promises and pitfalls. In controlled experiments on three algorithmic tasks, composed supervision progressively extends reliable prediction beyond the model's initial range, and each bottleneck appears clearly: weaker starting models gain less, invalid composition rules cause improvement to break down, and filtering restores progress but can leave coverage gaps (Section~\ref{sec:setup}). Beyond controlled tasks, composed supervision also outperforms direct self-improvement in program generation and multi-hop reasoning, although unreliable component predictions limit further gains, and incomplete supervision coverage leads to weaker generalization on unseen compositions (Section~\ref{sec:programming_reasoning}).
Finally, we show that even imperfect composed supervision can provide a useful warm start for reinforcement learning (Section~\ref{sec:rl_warm_start}).

Our contributions are as follows:
\begin{enumerate}[leftmargin=*,noitemsep,topsep=2pt]
\item We develop a framework for analyzing compositional self-improvement, relating the reliability of composed supervision to component errors, composition errors, and supervision coverage.

\item We identify three bottlenecks to continued improvement---{\emph{unreliable component predictions}}, \emph{structural errors in composition}, and \emph{limited supervision coverage}---and use controlled experiments to systematically study their effects and the limitations of filtering.

\item We show that composed supervision enables models to improve beyond the problems they can initially solve reliably. Even when the identified failure modes limit the extent of self-improvement, composed supervision can still provide a useful warm start for reinforcement learning.

\end{enumerate}

\section{Related Work}\label{sec:related_work}

\noindent\textbf{Compositional generalization.} Compositional generalization concerns handling novel combinations of familiar components \citep{lake_baroni_2018_scan,keysers_2020_cfq,kim_linzen_2020_cogs}. Prior work studies the roles of architectural biases, training scale, and data coverage \citep{bahdanau_2019_systematic,redhardt_2025_scaling}. Least-to-most and decomposed prompting use simpler subproblems to improve predictions at inference time \citep{zhou_2023_least_to_most,khot_2023_decomposed_prompting}. We examine whether training on composed targets improves direct prediction, including generalization to longer compositions and to cases not directly covered by the composed supervision.

\noindent\textbf{Synthetic data generation using compositionality.} Compositional data augmentation constructs new training examples by recombining existing examples or their constituent parts \citep{jia_liang_2016_recombination,andreas_2020_geca, akyurek_2021_recombine}. \citet{qiu_2022_compositional_augmentation} further show that grammar-generated training data can improve predictions beyond the generator's coverage. Recent work similarly composes existing problems into harder instances for curriculum reinforcement learning \citep{motwani_2025_h1,xu_2026_composition_rl}. A closely related use of compositional structure appears in iterated amplification, which constructs supervised targets for harder problems through decomposition and model-generated subanswers \citep{christiano_2018_amplification}. Our work studies a related question in the setting of self-improvement: when predictions on simpler components can provide reliable supervision for harder problems, and what limits the resulting capability expansion.

\noindent\textbf{Self-improvement.} Models can improve by training on self-generated solutions or feedback \citep{zelikman_2022_star,huang_2023_llms_self_improve,singh_2024_rest_em}. Iterative self-training can extend performance to harder algorithmic instances, although errors in self-generated labels can propagate across rounds \citep{lee_2025_self_improving}. System~2 distillation trains a model on answers produced by its own multi-step inference procedures to improve direct prediction \citep{yu_2024_system2_distillation}. Our study examines how component errors, composition errors, and supervision coverage constrain repeated training on composed predictions. Complementing work on learning compositions of familiar operations through RL \citep{yuan_2025_rl_compositionality}, we also test composed supervision as an RL warm start.
\section{Understanding Compositional Self-Improvement}
\label{sec:setup}
We first formalize compositional self-improvement and its training procedure. We then use controlled algorithmic tasks to understand when it works and why it can fail. These tasks let us separately vary component reliability, composition validity, and supervision coverage.
After showing that composed supervision can extend reliable prediction beyond the initial range, we examine how each of these factors can limit further improvement. We provide theoretical characterizations of these failure modes in \Cref{app:theory_proofs}.

\begin{figure*}[t!]
\centering
\includegraphics[width=0.98\textwidth]{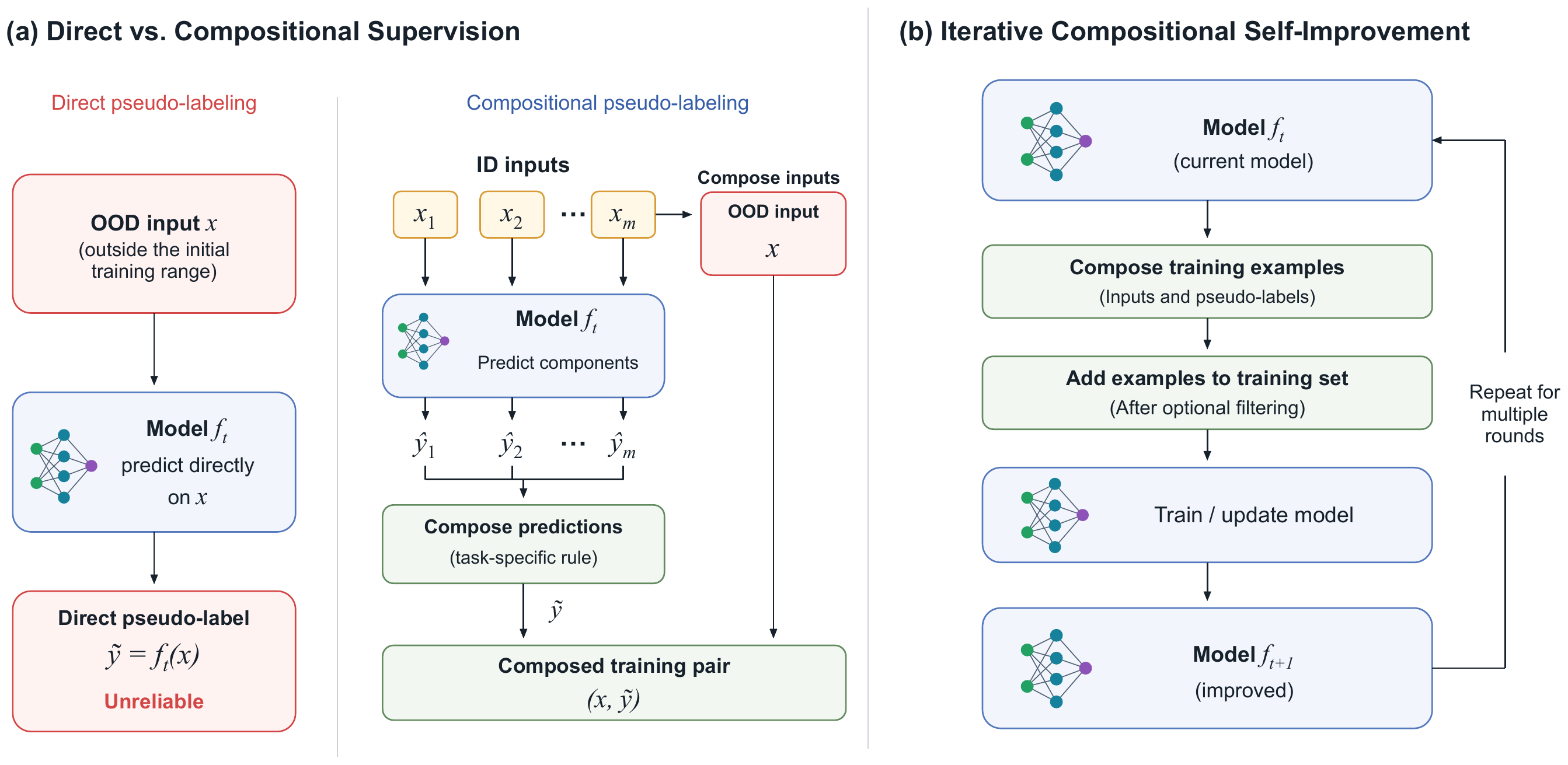}
\caption{Overview of compositional self-improvement. Direct pseudo-labeling can be unreliable on harder instances. Compositional self-improvement constructs targets by combining predictions on simpler subproblems, filters the resulting examples, and trains the model to solve larger problems directly. Updated predictions provide supervision for subsequent rounds.}
\label{fig:main_intuition}
\end{figure*}

\subsection{Compositional Self-Improvement}
\label{subsec:csi_method}

\begin{wrapfigure}[17]{R}{0.54\textwidth}
\vspace{-6mm}
\begin{minipage}{\linewidth}
\small
\begin{algorithm}[H]
\caption{Compositional self-improvement}
\label{alg:comp_self_improvement}
\begin{algorithmic}[1]
\REQUIRE initial model $\hat f_0$, schedule $\mathcal S$,
rule $A$, filter $G$, rounds $T$
\FOR{$t=1,\ldots,T$}
    \STATE $\widetilde D_t \gets \emptyset$
    \FORALL{$(x,\pi)$ selected by $\mathcal S$ at round $t$}
        \STATE obtain $\hat y_1,\ldots,\hat y_m$ by following
        $\pi$ with $\hat f_{t-1}$
        \STATE $\tilde y \gets A(\pi,\hat y_1,\ldots,\hat y_m)$
        \STATE add $(x,\tilde y)$ to $\widetilde D_t$
        if accepted by $G$
    \ENDFOR
    \STATE train $\hat f_t$ on $\widetilde D_t$ with
    task-specific replay
\ENDFOR
\STATE \textbf{return} $\hat f_T$
\end{algorithmic}
\end{algorithm}
\end{minipage}
\vspace{-4mm}
\end{wrapfigure}
Let $f^\star:\mathcal X\to\mathcal Y$ be the unknown target function and $\hat f_0$ an initial model trained on simpler problems.
For the controlled tasks below, a task-specific level $\ell:\mathcal X\to\mathbb N$ indexes problem scale, such as string length or the number of rooms. This level describes the task, not the model's accuracy. Compositional self-improvement uses predictions on simpler components to construct supervision for higher-level problems (\Cref{fig:main_intuition}).

For a parent problem $x$, a supplied decomposition $\pi$ specifies how to obtain its subproblems $x_1,\ldots,x_m$, including their order and any dependencies on earlier answers. A composition rule $A$ combines their answers. We say the composition is valid if using correct answers at every step produces the correct answer to the parent:
\begin{equation}
\label{eq}
A\bigl(\pi,f^\star(x_1),\ldots,f^\star(x_m)\bigr)=f^\star(x).
\end{equation}


Parent problems can be constructed from available components or paired with supplied decompositions. At round $t$, the model follows $\pi$ to obtain predictions $\hat y_1,\ldots,\hat y_m$, using earlier predictions in later queries when required. The candidate target is $\tilde y=A(\pi,\hat y_1,\ldots,\hat y_m)$, where aggregation may be performed by a fixed program or a model. An optional filter $G$ retains examples based on their structure or predicted answers, without consulting reference answers for the parent problems.

The model trains on retained examples and repeats with its updated predictions (\Cref{alg:comp_self_improvement}).

\subsection{Experimental Setup}
\label{subsec:controlled_setup}

We first study compositional self-improvement on three algorithmic tasks where the composition procedure and sources of error can be controlled. These tasks let us test whether composed supervision expands the range of problems models can reliably solve and separately examine component accuracy, composition errors, and supervision coverage.

\definecolor{csiA}{HTML}{1F6F8B} \definecolor{csiB}{HTML}{7B4FA8} \definecolor{csiR}{HTML}{E07A3F} \newcommand{\cA}[1]{\textcolor{csiA}{#1}} \newcommand{\cB}[1]{\textcolor{csiB}{#1}} \newcommand{\cR}[1]{\textcolor{csiR}{#1}} \newcommand{\rlstr}[1]{\texttt{\bfseries #1}} \newcommand{\gridtxt}[1]{{\scriptsize\ttfamily\bfseries\renewcommand{\arraystretch}{0.85}#1}}

\begin{table*}[t]
\centering
\caption{Composition examples for the controlled tasks. Grid routes are joined by the moves through the corridor between rooms, addition concatenates block sums, and max run length keeps the longer run. Filters reject carries across blocks and runs across the join. Blue and purple mark the components; orange marks parts added by the rule.}
\label{tab:controlled_composition_examples}
\small \renewcommand{\arraystretch}{1.35} \setlength{\tabcolsep}{4pt}
\begin{tabular}{@{}>{\raggedright\arraybackslash}p{0.16\textwidth}
                >{\raggedright\arraybackslash}p{0.185\textwidth}
                >{\raggedright\arraybackslash}p{0.185\textwidth}
                >{\raggedright\arraybackslash}p{0.25\textwidth}
                >{\raggedright\arraybackslash}p{0.14\textwidth}@{}}
\toprule
Task & Component 1 & Component 2 & Composed parent $\to$ label & Filter \\
\midrule
Grid navigation
& \gridtxt{\begin{tabular}[t]{@{}l@{}}\cA{\#\#\#\#\#\#\#}\\\cA{\#.....\#}\\\cA{\#...\#.\#}\\\cA{\#.....\#}\\\cA{\#.\#..G\#}\\\cA{\#S...\#\#}\\\cA{\#\#\#\#\#\#\#}\end{tabular}}\newline $\to$ \rlstr{\cA{RRURR}}
& \gridtxt{\begin{tabular}[t]{@{}l@{}}\cB{\#\#\#\#\#\#\#}\\\cB{\#....G\#}\\\cB{\#S....\#}\\\cB{\#\#\#...\#}\\\cB{\#\#....\#}\\\cB{\#.....\#}\\\cB{\#\#\#\#\#\#\#}\end{tabular}}\newline $\to$ \rlstr{\cB{URRRR}}
& \gridtxt{\begin{tabular}[t]{@{}l@{}}\#\cA{\#\#\#\#\#}\cR{\#\#\#}\cB{\#\#\#\#\#}\#\\\#\cA{.....}\cR{\#\#\#}\cB{....G}\#\\\#\cA{...\#.}\cR{\#..}\cB{.....}\#\\\#\cA{.....}\cR{\#.\#}\cB{\#\#...}\#\\\#\cA{.\#...}\cR{..\#}\cB{\#....}\#\\\#\cA{S...\#}\cR{\#\#\#}\cB{.....}\#\\\#\cA{\#\#\#\#\#}\cR{\#\#\#}\cB{\#\#\#\#\#}\#\end{tabular}}\newline $\to$ \rlstr{\cA{RRURR}\cR{RRUURR}\cB{URRRR}}
& No filter \\
\midrule
Addition
& $\cA{11}+\cA{22}=\cA{33}$
& $\cB{33}+\cB{44}=\cB{77}$
& $\cA{11}\cB{33}+\cA{22}\cB{44}=\cA{33}\cB{77}$
& Accept \\
& $\cA{11}+\cA{22}=\cA{33}$
& $\cB{60}+\cB{45}=\cB{105}$
& $\cA{11}\cB{60}+\cA{22}\cB{45}=\cA{33}\cB{105}$
& Reject \\
\midrule
Max run length
& \rlstr{\cA{3111}} $\to$ \rlstr{\cA{1|3}}
& \rlstr{\cB{022227}} $\to$ \rlstr{\cB{2|4}}
& \rlstr{\cA{3111}\cB{022227}} $\to$ \rlstr{\cB{2|4}}
& Accept \\
{\footnotesize (symbol\texttt{|}length of the longest run)}
& \rlstr{\cA{01111}} $\to$ \rlstr{\cA{1|4}}
& \rlstr{\cB{11200}} $\to$ \rlstr{\cB{1|2}}
& \rlstr{\cA{01111}\cB{11200}} $\to$ \rlstr{\cA{1|4}}
& Reject \\
\bottomrule
\end{tabular}
\end{table*}

\paragraph{Tasks.}
We consider three controlled tasks with different composition rules (\Cref{tab:controlled_composition_examples}).
\emph{Grid navigation} requires predicting a route from a start to a goal in a grid. We construct larger grids by connecting smaller rooms with corridors and compose their predicted routes by inserting the corridor moves between them.
\emph{Addition} requires predicting the sum of two integers. We construct longer operands by concatenating smaller blocks and compose their predicted sums by concatenation, filtering cases where carries cross block boundaries.
\emph{Max run length} requires identifying the symbol and length of the longest run of repeated symbols in a string. We concatenate two strings and select the longer of their predicted runs, filtering cases where a longer run crosses the concatenation boundary.

For each task, the difficulty level corresponds to the number of rooms, the maximum addend length, and the string length, respectively, with seed ranges of one room, $3$--$7$ digits, and lengths $6$--$10$.
The self-improvement schedules progressively extend these ranges: grid navigation adds one room per round for 12 rounds (up to 13 rooms), addition adds two digits per round for 8 rounds (up to 23 digits), and max run length first extends to length 20, then adds nine symbols per round for 8 rounds total (up to length 83). Each round generates composed examples across all expanded levels up to the current maximum (\Cref{appsubsec:training_setup}).

\paragraph{Models and training.}
All three tasks use 14.2M-parameter character-level Transformers. For each task, we first train a model on the initial difficulty range and then construct supervision for progressively harder instances from predictions on simpler components. 
Each round trains on these composed examples alongside replay at the initial training levels to help preserve previously learned capabilities. We evaluate held-out instances across both the seed and expanded difficulty ranges after each round. For all three tasks, we additionally compare against direct self-improvement, which uses the model's prediction on the full harder instance as its pseudo-label under the same expansion schedule (\Cref{appsubsec:algorithmic_baselines}). Full training details appear in \Cref{appsubsec:training_setup}.

\subsection{Composed Supervision Expands What Models Can Solve}
\label{subsec:main_self_improvement}
\begin{figure*}[t!]
\centering
\includegraphics[width=\textwidth]{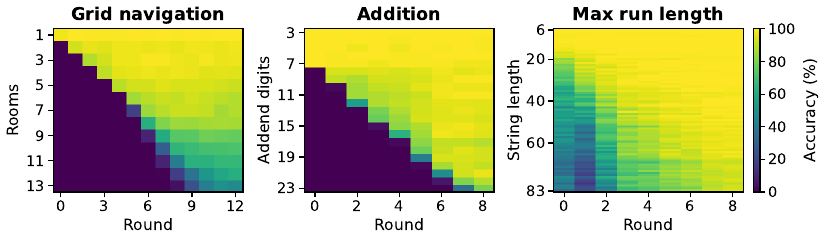}
\caption{Compositional self-improvement expands the range of instances solved by the model. Each heatmap shows accuracy across self-improvement rounds. From left to right: grid navigation by the number of rooms, addition by addend digit length, and max run length by string length. In all three tasks, reliable prediction extends beyond the initial model's reliable range.}
\label{fig:self_improvement_panels}
\vspace{-3mm}
\end{figure*}

\Cref{fig:self_improvement_panels} shows that compositional self-improvement extends reliable prediction beyond the initial range in all three tasks, while direct self-improvement under the same schedules yields little or no such improvement on addition and grid navigation (\Cref{appsubsec:algorithmic_baselines}). Predictions on simpler instances can thus supervise instances beyond the model's reliable range. At the supervision level, our analysis makes this advantage precise: composed labels are more accurate than direct self-labels whenever the combined component and composition errors fall below the model's direct error on the same problems, a simple sufficient condition formalized in \cref{eq:csi_supervision_advantage}.

\subsection{Failure Modes of Compositional Self-Improvement}
\label{subsec:failure_modes}

While \Cref{subsec:main_self_improvement} shows that compositional self-improvement yields consistent gains, these gains are based on three conditions: reliable component predictions, valid composition, and supervision that covers the harder instances. We illustrate how the gains are limited when each condition fails: weaker starting models plateau early in grid navigation, invalid compositions stall progress in addition unless filtered, and in max run length, composed labels never contain runs longer than those of their components, and accuracy on such runs falls sharply.

\begin{figure}[!t]
\centering
\includegraphics[width=\textwidth]{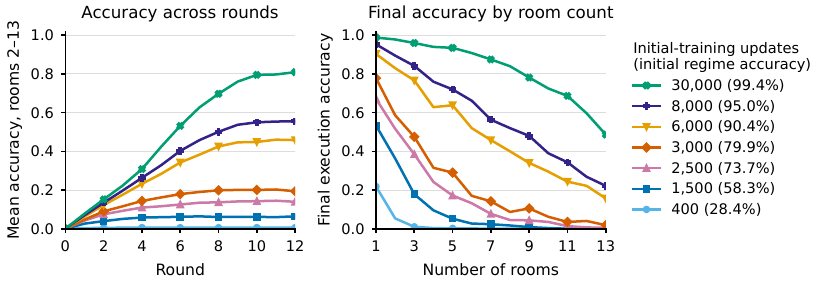}
\caption{Weaker starting models can limit compositional self-improvement, as illustrated by experiments on the grid navigation task. Left: mean accuracy on 2--13 rooms across rounds. Right: accuracy by room count after the final round. The legend gives initial-training updates and one-room accuracy. Curves average three seeds.}
\label{fig:grid_seed_accuracy}
\end{figure}

\paragraph{Failure mode I: Low component accuracy.}
Compositional self-improvement relies on accurate component predictions to construct reliable supervision for larger instances. We examine this failure mode in grid navigation by varying the initial model's accuracy and applying the same twelve-round schedule to each starting model (\Cref{appsubsec:grid_seed_accuracy}). Weaker starting models plateau earlier, while stronger models sustain larger gains (\Cref{fig:grid_seed_accuracy}). Models with initial one-room accuracies of $79.9\%$ and $99.4\%$ reach final mean accuracies of $19.4\%$ and $80.9\%$ on $2$--$13$ rooms, respectively. \textbf{Reliable component predictions are therefore crucial for gains by compositional self-improvement}. Our analysis captures this directly: under a valid composition rule, component errors pass directly into the composed label (\Cref{prop:csi_component_obstruction}).

\begin{figure}[!t]
\centering
\includegraphics[width=\textwidth]{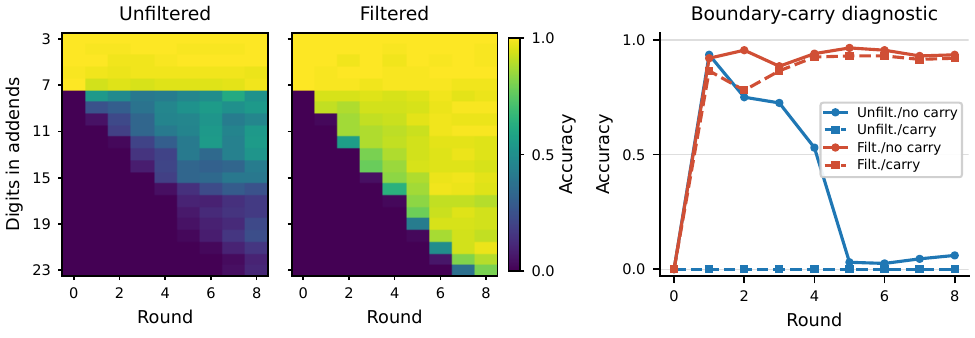}
\caption{Structural errors can undermine self-improvement, although filtering can restore gains in the ``addition'' tasks including on test sums with boundary carries. Left and middle: accuracy by addend length across rounds without and with filtering. Right: accuracy at each round's newest lengths (up to $22$--$23$ digits), on sums with and without a carry across the block boundary.}
\label{fig:addition_filtering}
\end{figure}

\paragraph{Failure mode II: Structural errors from invalid composition rules.}
\label{subsec:addition_baselines}
In practice, composition rules designed for self-improvement may be imperfect, introducing errors even when all component predictions are correct. We illustrate this with addition: concatenating blockwise sums can produce incorrect labels when a carry crosses the block boundary. When such failure conditions are known, we can filter out invalid compositions. This removes incorrect labels but also every parent with a carry across its own block boundary. \Cref{fig:addition_filtering} shows that unfiltered composition degrades after initial gains, while filtering restores self-improvement. The filtered model still solves test sums with boundary carries (\Cref{fig:addition_filtering}, right). Since filtering still permits correct carries within individual blocks, this may reflect generalization of learned carry operations to different digit positions.



\begin{wrapfigure}{r}{0.36\textwidth}
\centering
\includegraphics[width=\linewidth]{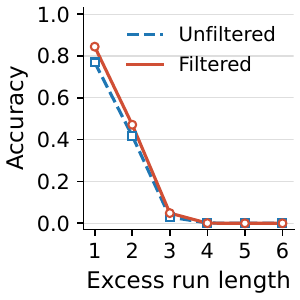}
\captionsetup{font=small}
\caption{Boundary-crossing accuracy in the max run length task.
Excess run length is the difference between the longest run
in the concatenated string and the longest run in either component.}
\label{fig:run_length_boundary_support}
\vspace{-8mm} 
\end{wrapfigure}

\paragraph{Failure mode III: Limited supervision coverage.}
Even when composed labels are correct, they may cover only some of the output patterns that arise in harder instances. In max run length, for example, if we build ten-symbol strings only from pairs of five-symbol strings, the rule can correctly supervise runs of at most five symbols: concatenating \texttt{01111} and \texttt{11200} requires \texttt{1|6}, but the rule produces \texttt{1|4}. Filtering removes such errors but leaves longer runs without correct supervision. Thus, although self-improvement achieves high overall max run length accuracy (\Cref{fig:self_improvement_panels}), this aggregate success conceals poor performance on runs beyond the coverage of composed supervision. \Cref{fig:run_length_boundary_support} breaks down accuracy by excess run length, the amount by which a boundary-crossing run exceeds the longest run in either component. Accuracy drops sharply as this excess increases, approaching zero at an excess of three symbols, both with and without filtering.

Our analysis formalizes this limitation: if composition stays within a restricted output space, valid composition cannot directly supervise outputs outside it (\Cref{prop:csi_output_coverage}). Unlike the boundary carries in addition, these runs require outputs that no composed label contains, and the model does not extrapolate to them.
\section{Compositional Self-Improvement Beyond Controlled Tasks}
\label{sec:programming_reasoning}
Next, we analyze compositional self-improvement on two real-world tasks with natural compositional structure: program generation and multi-hop reasoning. In both, the compositional and direct methods label the same parent problems under the same schedule (\Cref{tab:realworld_composition_examples}).

\begin{table*}[t]
\centering
\caption{Composition examples for program generation and multi-hop reasoning.}
\label{tab:realworld_composition_examples}
\small \renewcommand{\arraystretch}{1.2} \setlength{\tabcolsep}{4pt}
\begin{tabular}{@{}>{\raggedright\arraybackslash}p{0.10\textwidth}
                >{\raggedright\arraybackslash}p{0.18\textwidth}
                >{\raggedright\arraybackslash}p{0.18\textwidth}
                >{\raggedright\arraybackslash}p{0.30\textwidth}
                >{\raggedright\arraybackslash}p{0.16\textwidth}@{}}
\toprule
Task & Component 1 & Component 2 & Composed parent $\to$ label & Filter \\
\midrule
Program gen\-eration
& Output the first 5 lines $\to$ \rlstr{\cA{head -5}}
& Sort in descending order $\to$ \rlstr{\cB{sort -r}}
& Step 1: output the first 5 lines. Step 2: sort in descending order.
  $\to$ \rlstr{\cA{head -5}\cR{\ |\ }\cB{sort -r}}
& Parses as a pipeline \\
\midrule
Multi-hop reasoning
& Who issued the Royal Proclamation of 1763? $\to$ \cA{King George III}
& What year did \cA{King George III} die? $\to$ \cB{1820}
& What year did the person who issued the Royal Proclamation of 1763 die?
  $\to$ \cB{1820}
& Answer format; answer found in the passages \\
\bottomrule
\end{tabular}
\end{table*}

\subsection{Program generation}
\label{subsec:extension_sft}

Many programs are assembled from smaller programs, making program generation a natural target for composition. We study Bash pipeline generation, where the model takes a natural-language description of a sequence of text-processing steps and generates a pipeline implementing them. Compositional self-improvement constructs supervision for the full pipeline by joining commands predicted for individual steps using the pipe operator ($|$), which passes the output of one command to the next  (\Cref{tab:realworld_composition_examples}).

\begin{figure}[!htb]
\centering
\includegraphics[width=\linewidth]{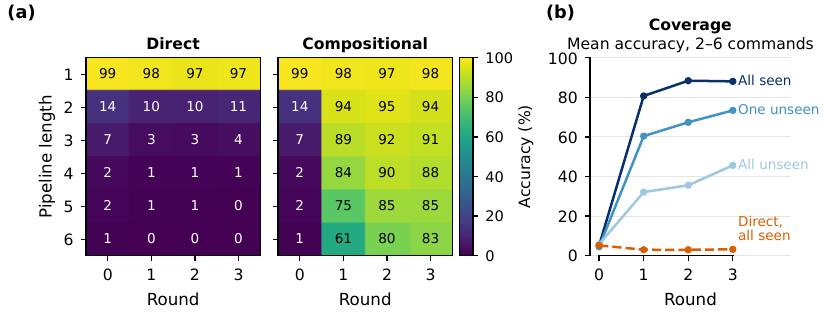}
\caption{Compositional self-improvement improves program generation, with less transfer to commands absent from training compositions. The initial model is trained on all individual commands; a command is \emph{seen} if it also appears in training compositions. (a) Accuracy of direct and compositional self-improvement on test pipelines of seen commands; length 1 tests single commands. (b) Mean accuracy on 2- to 6-command test pipelines whose commands are all seen, include one unseen command, or are all unseen. Round 0 is the initial model; round $r$ trains on pipelines of length $r+1$.}
\label{fig:program_tasks_sft}
\end{figure}

\paragraph{Setup.}
We build pipelines from 119 command variants in TLDR\footnote{\url{https://github.com/tldr-pages/tldr}} and NL2Bash~\citep{lin_2018_nl2bash}. Both methods start from Qwen3.5-2B trained on all 119 individual commands, rehearse them every round, and run three rounds on two-, three-, and four-command pipelines. Compositional self-improvement joins its predicted commands for the individual steps with pipes, while direct self-improvement predicts the full pipeline; both keep a label only if it parses. We evaluate generated pipelines by execution, averaging four seeds. Details are in \Cref{appsubsec:bash_setup}.


\paragraph{Results.}
\Cref{fig:program_tasks_sft}(a) shows that composed supervision substantially improves program generation on pipelines using commands seen in training compositions, including five- and six-command pipelines that are longer than those used for training. Direct self-improvement makes little progress.

Additionally, to examine the effects of limited coverage in composed supervision, we evaluate the same models on pipelines containing 30 commands excluded from multi-command training but retained in initial single-command training and replay. After three rounds, mean accuracy on two- to six-command pipelines drops from 88\% when all commands appear in training compositions to 73\% with one held-out command and 46\% when all are held out (\Cref{fig:program_tasks_sft} (b)). This shows that limited supervision coverage can constrain generalization even when individual commands are learned.


\subsection{Multi-hop reasoning}
\label{subsec:reasoning_sft}

Reasoning often chains simple steps, each using the last one's result. We study this with multi-hop question answering, using the decompositions annotated in MuSiQue. Unlike in program generation, later steps depend on earlier predictions: the model answers the sub-questions in order, substituting each answer into the next, and its final answer becomes the label.

\begin{figure}[!htb]
\centering
\includegraphics[width=\linewidth]{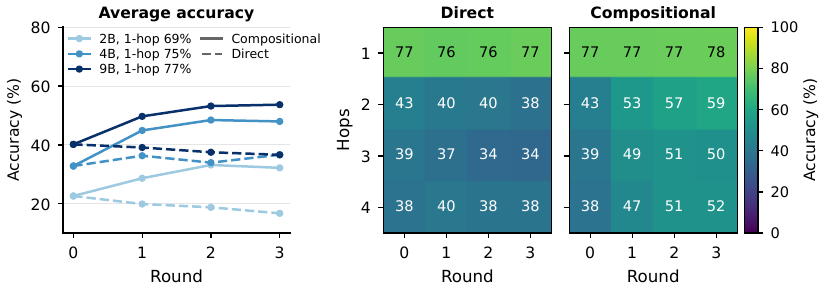}
\caption{Composed supervision improves multi-hop reasoning on MuSiQue. Left: mean accuracy on 2- to 4-hop questions for all sizes, labeled with each initial model's single-hop accuracy. Middle and right: exact-match accuracy of Qwen3.5-9B by hop count under direct and compositional self-improvement. Round 0 is the initial model, trained on single-hop questions; rounds 1--3 train on 2-, 3-, and 4-hop questions.}
\label{fig:musique_hop_ladder}
\end{figure}

\paragraph{Setup.}
We use MuSiQue~\citep{csi_musique_2022}, which provides 2- to 4-hop questions with annotated decompositions, and train Qwen3.5-2B, 4B, and 9B to examine how the gains depend on model size. Each model is first trained on single-hop questions with gold answers and then runs three rounds on 2-, 3-, and 4-hop questions, respectively. The compositional method follows the annotated decomposition, while direct self-improvement answers the original question in one step. Filters check the answer format and whether the answer occurs in the passages or the question, without reference answers; the compositional method applies the same checks to its intermediate answers. At evaluation, the model answers each question directly from the provided passages, without decompositions. Results average three seeds; details are in \Cref{appsubsec:musique}.

\paragraph{Results.}
\Cref{fig:musique_hop_ladder} shows that composed supervision improves multi-hop reasoning at all three model sizes, while direct self-improvement makes little progress. The gains extend beyond the hop counts used for training: 4-hop accuracy rises during the first two rounds, which train only on 2- and 3-hop questions. Component reliability limits the gains, consistent with failure mode I. Errors in intermediate answers can propagate through composition, and first-round composed-label accuracy rises from 45\% for 2B to 65\% for 9B, with most errors arising at the final step. The 2B model, whose components are least reliable, gains least. Additional analyses of label errors appear in \Cref{appsubsec:label_accuracy}.

\begin{table*}[t]
\centering
\caption{Composition examples for the RL warm-start tasks, showing two functions of longer chains.}
\label{tab:rl_composition_examples}
\small \renewcommand{\arraystretch}{1.2} \setlength{\tabcolsep}{4pt}
\begin{tabular}{@{}>{\raggedright\arraybackslash}p{0.10\textwidth}
                >{\raggedright\arraybackslash}p{0.16\textwidth}
                >{\raggedright\arraybackslash}p{0.20\textwidth}
                >{\raggedright\arraybackslash}p{0.34\textwidth}
                >{\raggedright\arraybackslash}p{0.12\textwidth}@{}}
\toprule
Task & Component 1 & Component 2 & Composed parent $\to$ label & Filter \\
\midrule
String trans\-formations
& \rlstr{func\_20(kqz)}\newline $\to$ \rlstr{\cA{k-q-z}}
& \rlstr{func\_11(\cA{k-q-z})}\newline $\to$ \rlstr{\cB{-q-zk}}
& \rlstr{func\_11(func\_20(kqz))}\newline $\to$
  \rlstr{\cA{func\_20=k-q-z}\cR{\ |\ }\cB{func\_11=-q-zk}\cR{\ =>\ }\cB{-q-zk}}
& Each step nonempty, length $\leq 96$ \\
\midrule
Typed Python
& Last digit $\to$ $\cA{p_1}$: \rlstr{\cA{return x \% 10}}
& Integer to binary string $\to$ $\cB{p_2}$: \rlstr{\cB{return bin(x)[2:]}}
& Step 1: last digit. Step 2: integer to binary string. $\to$
  \rlstr{\cR{return }\cB{p2}\cR{(}\cA{p1}\cR{(x))}}
& Parses as one function \\
\bottomrule
\end{tabular}
\end{table*}

\begin{figure*}[t]
\centering
\includegraphics[width=0.8\textwidth]{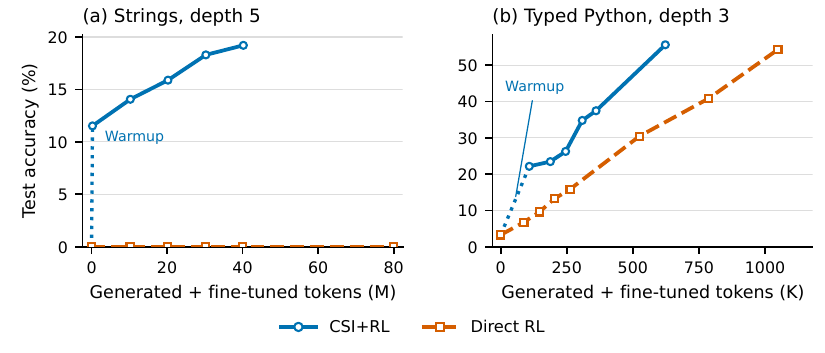}
\caption{A compositional SFT warmup improves subsequent RL on five-operation string transformations and three-function Python programs. CSI+RL first fine-tunes on labels from compositional self-improvement (CSI), then runs RL, using half the token budget of direct RL. Curves show mean test accuracy over three seeds; dotted segments mark the SFT warmup. Token counts include label generation, RL samples, and warmup targets.}
\label{fig:rl_frontier}
\end{figure*}

\section{Compositional Supervision as an RL Warm Start}
\label{sec:rl_warm_start}
Although the failure modes above limit compositional self-improvement on its own, it can still give RL a useful starting point. \citet{yuan_2025_rl_compositionality} show that RL can teach a model to compose existing skills, but learning can be extremely inefficient when correct samples are rare. We show that a short SFT warmup stage on composed targets, even imperfect ones, can accelerate RL.

\paragraph{Setup.}
We use string transformations from \citet{yuan_2025_rl_compositionality} and Python programs composed from type-compatible functions in MBPP+ and HumanEval+ \citep{csi_mbpp_2021,chen_2021_humaneval,csi_evalplus_2023}. In string transformations, the model takes an input string and a sequence of operations and predicts their intermediate and final outputs. In Python program generation, the model takes natural-language specifications of type-compatible functions and generates code that applies them in sequence.
\Cref{tab:rl_composition_examples} illustrates their composition rules: for string transformations, each predicted output becomes the next operation's input; for Python, separately generated functions are assembled into a wrapper that calls them in sequence.
For initial training, we train a six-layer Transformer on one- and two-operation chains for string transformations and Qwen3.5-2B on single-function problems for Python.
We then compare direct RL against RL following a short SFT warmup on composed targets, using five-operation chains and three-function Python programs. Both use GRPO \citep{shao_2024_deepseekmath} with binary outcome rewards. We compare test accuracy against token usage, counting warmup label generation, warmup target tokens, and all RL samples. Additional details are in \Cref{appsubsec:rl_curves,appsubsec:rl_setup}.



\paragraph{Results.}
\Cref{fig:rl_frontier} shows that \textbf{composed supervision provides an effective warm start for RL}. On five-operation string chains, direct RL never samples a correct composition and receives no reward, even with twice the generation budget, whereas RL after a short compositional SFT warmup improves steadily. On typed Python, where direct RL also makes progress, RL after the warmup reaches $55.6\%$ with half the generation budget, while direct RL needs the full budget to reach $54.3\%$. Here composition plays a different role from \Cref{sec:setup,sec:programming_reasoning}: the composed targets need not be the final training signal, only move the model to a regime where correct compositions are sampled often enough for outcome rewards to guide learning. Nor must they be fully correct; only $77\%$ of the string warmup labels are. More results appear in \Cref{appsubsec:rl_curves,appsubsec:label_accuracy}.

\ifdefined\section{Theoretical Analysis}
\label{sec:theory_main}

We provide a theoretical characterization of when compositional self-improvement succeeds and why it can fail. Our analysis proceeds as follows. First, we show how the reliability of composed supervision depends on errors in component predictions and the composition rule. Second, we use these results to characterize the three failure modes observed in our experiments: unreliable components, invalid composition, and limited supervision coverage. Finally, we establish a sufficient condition under which training on composed labels improves the model. Full statements and proofs appear in \Cref{app:theory_proofs}.

\paragraph{Setup.}
We follow the notation of \Cref{subsec:csi_method}. At difficulty level $k$, let $X\sim D_k$ be a candidate parent problem. The current model $\hat f$ predicts answers to simpler components, which are combined into a composed label $\widetilde Y$. An optional filter $G\in\{0,1\}$ determines whether the resulting example is retained for training.

To distinguish errors inherited from the model from errors introduced by composition, let $\widetilde Y^\circ$ denote the label obtained by executing the same composition procedure with correct component answers. This label need not equal the correct parent answer $f^\star(X)$ if the composition rule is invalid.

We define three error rates on the accepted examples:
\begin{equation}
\label{eq:main_error_rates}
\begin{aligned}
\rho_k
&= \Pr_k[\widetilde Y\ne f^\star(X)\mid G=1],\\
\epsilon_k
&= \Pr_k[E_{\mathrm{comp}}\mid G=1],\\
\xi_k
&= \Pr_k[\widetilde Y^\circ\ne f^\star(X)\mid G=1],
\end{aligned}
\end{equation}
where $E_{\mathrm{comp}}$ is the event that at least one component prediction is incorrect, and $\Pr_k$ denotes probability over the candidate-sampling and composition process. Here, $\rho_k$ measures the overall composed-label error, $\epsilon_k$ measures errors in the component predictions, and $\xi_k$ measures errors that remain even with correct components. All three quantities are evaluated on the same accepted examples ($G=1$).

\paragraph{When is composed supervision more reliable?}
The key observation is simple: a composed label can differ from $\widetilde Y^\circ$ only if at least one component prediction is incorrect. This gives the following bound.

\begin{proposition}[Reliability of composed labels]
\label[proposition]{prop:main_reliability}
The composed-label error satisfies
\begin{equation}
\label{eq:main_reliability}
|\rho_k-\xi_k|\le\epsilon_k,
\qquad
\rho_k\le\min\{1,\epsilon_k+\xi_k\}.
\end{equation}
If each execution makes at most $m$ component calls, then $\epsilon_k\le\sum_{i=1}^{m}\epsilon_{k,i}$, where $\epsilon_{k,i}$ is the probability, conditional on $G=1$, that call $i$ occurs and is answered incorrectly.
\end{proposition}

The bound does not require component errors to be independent, and it allows filtering decisions to depend on model predictions (\Cref{prop:k_component_noise}).

This result also provides a sufficient condition for composed supervision to outperform direct self-improvement. Let $\rho_k^{\mathrm{direct}} =\Pr_k[\hat f(X)\ne f^\star(X)\mid G=1]$ be the current model's error when answering the same accepted parent problems directly. Composed labels are more accurate whenever
\begin{equation}
\label{eq:main_supervision_advantage}
\epsilon_k+\xi_k<\rho_k^{\mathrm{direct}}.
\end{equation}
Thus, composed labels need not be perfect: their combined component and composition errors must simply be smaller than the model's direct prediction error on the harder problems.

\paragraph{Three failure modes.}
The reliability bound also helps explain when compositional self-improvement breaks down.

\emph{(I) Unreliable component predictions (\Cref{prop:csi_component_obstruction}).} Even with a valid composition rule, errors in individual components can accumulate. For fixed-query compositions in which component errors cannot be masked, we have $\rho_k=\epsilon_k$, with
\[
\max_i\epsilon_{k,i}
\le \rho_k
\le \min\Bigl\{1,\sum_{i=1}^{m}\epsilon_{k,i}\Bigr\}
\]
. The resulting error depends not only on component accuracy but also on how component errors overlap. Consequently, composing more components can make labels unreliable even when each component is relatively accurate. This helps explain why weaker starting models plateau earlier in grid navigation (\Cref{fig:grid_seed_accuracy}) and why component reliability limits gains in multi-hop reasoning.

\emph{(II) Invalid composition rules.} Improving component accuracy cannot eliminate errors introduced by the composition rule. In particular, \Cref{prop:main_reliability} gives $\rho_k\ge\xi_k-\epsilon_k$, and hence $\rho_k=\xi_k$ when the components are perfectly accurate. Addressing this failure requires correcting the rule or excluding invalid compositions. For example, filtering addition problems with carries across block boundaries removes a source of structural errors (\Cref{fig:addition_filtering}). However, such filtering also reduces the set of examples available for training.

\emph{(III) Limited supervision coverage (\Cref{prop:csi_output_coverage}).} Even perfectly accurate labels may fail to cover the outputs needed for harder problems. Consider a set of outputs $C\subseteq\mathcal Y$ that is closed under composition: if all correct component answers belong to $C$, the composed answer also belongs to $C$. For deterministic compositions with fixed component queries, let $p_{k,C}$ be the fraction of accepted parent problems whose correct answers lie outside $C$. Then
\[
p_{k,C}
\le
\xi_k+
\sum_{i=1}^{m}
\Pr_k[f^\star(X_i)\notin C\mid G=1]
\]

When the rule is valid on accepted examples ($\xi_k=0$), outputs outside $C$ can receive supervision only if they already occur among the correct component answers. This explains the max run length example: selecting the longer of two component runs cannot produce a run longer than either component, while filtering boundary-crossing runs removes the corresponding invalid labels (\Cref{fig:run_length_boundary_support}).

Moreover, when such outputs are rare in the accepted data, a model that never predicts them can still achieve nearly optimal pseudo-label accuracy. Under realizability and a closure assumption on the hypothesis class, the additional training error from restricting predictions to $C$ is exactly $p_{k,C}$ (\Cref{prop:csi_absent_output_penalty}). Thus, accurate composed supervision alone does not guarantee coverage of the patterns needed for further improvement.

\paragraph{When does reliable supervision improve the model?}
The preceding results concern the quality of composed labels. We now ask whether training on these labels actually improves prediction accuracy.

Let $R_k^+(h)=\Pr_k[h(X)\ne f^\star(X)\mid G=1]$ denote the true-label error of a predictor $h$ on accepted examples. Suppose a finite hypothesis class $\mathcal H$ contains $f^\star$, and the updated model $\hat f_{\mathrm{new}}\in\mathcal H$ approximately minimizes empirical pseudo-label error over $n$ independent accepted examples, up to optimization error $\tau\ge0$. Then, with probability at least $1-\delta$,
\begin{equation}
\label{eq:main_learning_bound}
R_k^+(\hat f_{\mathrm{new}})
\le
2\rho_k+
2\sqrt{\frac{\log(2|\mathcal H|/\delta)}{2n}}
+\tau
\end{equation}
(\Cref{thm:finite_class_noisy_erm}).

The bound separates three factors: composed-label error, finite-sample estimation error, and optimization error. Combined with \Cref{prop:main_reliability}, it guarantees improvement over the current model on accepted examples whenever
\[
2(\epsilon_k+\xi_k)
+2\sqrt{\frac{\log(2|\mathcal H|/\delta)}{2n}}
+\tau
<R_k^+(\hat f).
\]
Together, these results characterize the conditions for compositional self-improvement: reliable component predictions and valid composition rules determine the quality of composed supervision, while coverage determines which problems receive that supervision. When the resulting labels are sufficiently reliable and learnable, they can support further model improvement.\fi
\section{Conclusion}
We studied the promises and pitfalls of self-improvement that exploits the compositional structure of tasks, and our analysis yields three practical lessons. First, teach atomic skills well: component errors can propagate into composed labels and limit further improvement. Second, find where the composition rule breaks: even correct components can yield incorrect labels, so identifying and filtering invalid compositions is crucial. Third, check which outputs composed labels can reach: some rules restrict the outputs they can generate, leaving coverage gaps that more rounds or stricter filtering cannot fill.

Our study focuses on tasks with supplied decomposition plans and composition rules, leaving open how to discover and validate such structures in less controlled settings. Another important direction is to expand supervision coverage without introducing additional composition errors. Our results also suggest a complementary approach: even when the identified failure modes limit self-improvement on its own, composed supervision can provide a useful warm start for reinforcement learning. More broadly, these findings highlight the potential of task structure as a source of supervision, while emphasizing the importance of understanding its reliability and coverage.


\subsection*{AI Use Statement}

We used generative AI assistants (Claude and ChatGPT) extensively throughout this project.

\noindent\textbf{Not used for.} AI tools were not used to originate the central research question or hypotheses. The authors determined the research direction and made the final decisions on experimental design, analysis, interpretation, and the claims presented in this work.

\noindent\textbf{Used for.} We used AI assistants for the following tasks:
\begin{itemize}[leftmargin=*,noitemsep,topsep=2pt]
\item \emph{Software implementation and experiment execution:} Developing and debugging training, evaluation, data-construction, and plotting code, and preparing and running experiments on GPU clusters.
\item \emph{Experimental design and analysis:} Suggesting baselines, ablations, diagnostic analyses, experimental parameters, and possible interpretations of results.
\item \emph{Theoretical analysis:} Assisting with mathematical formulations, proof development and verification, and Lean~4 formalization.
\item \emph{Manuscript preparation:} Assisting with identifying relevant prior work, preparing figures, and polishing manuscript text.
\end{itemize}

The authors reviewed AI-assisted implementations and checked reported results against saved experimental outputs. The theoretical arguments were developed and reviewed by the authors, with AI assistance in refining mathematical formulations, checking proofs, and formalizing claims in Lean~4 for machine verification. Cited references were manually verified against their sources. All AI-assisted suggestions and outputs were subject to author review, and the authors take full responsibility for the final content of this work.

\label{csi-realworld-main-end}
\bibliographystyle{iclr2027_conference}
\bibliography{related_work}

\clearpage
\appendix
\clearpage
\section*{Appendix}
\Cref{appsec:experiment_details} defines the tasks and gives the experimental
setups, \Cref{app:theory_proofs} provides the theoretical analysis, and
\Cref{appsec:additional_experiments} reports additional results and analyses.

\section{Experimental Details}\label{appsec:experiment_details}

\subsection{Task Definitions}\label{appsubsec:comp_rule}
For each task, we give the task itself (input, target, and how correctness is scored), the composition rule that builds a parent's label from the model's answers to its components, the filter, and the data. Grid navigation, program generation, string transformations, and typed Python compose exactly; addition and max run length need a filter to exclude parents on which the rule fails. \Cref{tab:controlled_composition_examples,tab:realworld_composition_examples,tab:rl_composition_examples} show an example of each rule, and \Cref{fig:algorithmic_composition_example,fig:bash_composition_example,fig:musique_composition_example,fig:strings_composition_example,fig:python_composition_example} show the full model inputs and saved outputs for one training label per task.

\subsubsection{Grid navigation}
\paragraph{Task.} An instance is a chain of $k$ rooms drawn as one text grid (\texttt{\#} wall, \texttt{.} free, \texttt{S} start, \texttt{G} goal), and the model outputs a route over $\{\texttt{U},\texttt{D},\texttt{L},\texttt{R}\}$. Many routes reach the goal, so a route is correct if executing it from the start reaches the goal within a step cap set by the grid size. \citet{lee_2025_self_improving} self-improve on maze shortest paths given as adjacency lists; the room structure is introduced here and released as \textsc{ModularGrid}.

\paragraph{Composition.} Adjacent rooms are joined by a corridor of three columns, open only at the left room's exit row, across the rows between the exit and the entry, and at the right room's entry row. Every route must therefore leave each room by its exit and enter the next by its entry, and the corridor actions $b_i$ between rooms $i$ and $i+1$ are fixed by these two rows: \texttt{RR}, then \texttt{U} or \texttt{D} once per row between them, then \texttt{RR}. The shortest route of a chain is the concatenation of its rooms' shortest routes and the corridor actions,
\[
f^\star_{\mathrm{grid}}(x_1\circ\cdots\circ x_k) =f^\star_{\mathrm{grid}}(x_1)\Vert b_1\Vert\cdots\Vert b_{k-1}\Vert f^\star_{\mathrm{grid}}(x_k).
\]

\paragraph{Filter.} None: the corridors make the rule valid on every chain.

\paragraph{Data.} Each room has a $5\times5$ interior with wall probability $0.28$ and a shortest entry-to-exit path of $4$--$12$ steps. Initial training uses $200{,}000$ one-room grids. Each round builds $15{,}000$ new chains per room count, for every count from $2$ to the current maximum, from the replay rooms. The test set contains $200$ chains per room count from $1$ to $13$, generated before any training data. Test chains never appear in training, and their rooms are excluded from the composed pools, although many of them occur among the initial-training rooms, which come from the same generator.

\subsubsection{Addition}
\paragraph{Task.} The input is a pair of decimal integers $(A,B)$, and the target is their sum $A+B$.

\paragraph{Composition.} Both addends are split into aligned blocks, $A=a_1\Vert\cdots\Vert a_k$ and $B=b_1\Vert\cdots\Vert b_k$, where block $i$ has $d_i$ digits and $\Vert$ denotes concatenation. Each component is one block sum $s_i=a_i+b_i$, and the label concatenates the leading sum $s_1$, which may overflow its width, with the lower sums padded to their widths.

\paragraph{Filter.} The label is correct exactly when no carry crosses a block boundary, that is, $s_i<10^{d_i}$ for every non-leading block $i=2,\ldots,k$. The filter accepts only such parents; without it, a boundary carry gives a wrong label.

\paragraph{Data.} Initial training uses addends of $3$--$7$ digits (\Cref{tab:training_setup}). Each round samples $10{,}000$ new composed problems per addend length, for every length from $8$ to the current maximum, and $5{,}000$ new replay problems per seed length, labeled by the current model. The test set contains $100$ problems per length from $3$ to $23$, sampled uniformly and excluded from every training pool.

\subsubsection{Max run length}
\paragraph{Task.} The input is a string $x$ over the digits $0$--$9$. A run is a maximal block of equal symbols, and the target is the symbol and length of the longest run, written \texttt{symbol|length}; among runs of equal length, the leftmost counts.

\paragraph{Composition.} A parent is the concatenation $xy$ of two components, and its label is the answer of the component with the longer run, taking $x$ on ties because every run of $x$ precedes every run of $y$:
\[
f^\star_{\mathrm{run}}(xy)=
\begin{cases}
f^\star_{\mathrm{run}}(x), & L(x)\ge L(y),\\
f^\star_{\mathrm{run}}(y), & L(y)>L(x),
\end{cases}
\]
where $L$ is the length of the longest run.

\paragraph{Filter.} The rule fails when a run continues across the boundary, since the merged run can be longer than both component runs. The filter $G(x,y)=\mathbf 1\{x_{|x|}\neq y_1\}$ accepts a parent only if the last symbol of $x$ differs from the first symbol of $y$.

\paragraph{Data.} Initial training uses $50{,}000$ strings per length from $6$ to $10$, which are replayed with true labels in every round. Each round draws $2{,}000$ new composed strings per length, for every length from $12$ to the current maximum. The test set contains $100$ strings per length from $6$ to $83$, drawn before any training data and excluded from it. For the boundary sweep of \Cref{fig:run_length_boundary_support}\label{appsubsec:fixed_path_boundary_slices}, which tests inputs the filter excludes from training, we take a same-length reference string whose longest run has length $r$, build an input whose longest run has length $r+\Delta$ and crosses the midpoint, and evaluate the saved models of rounds $0$--$7$ on their current length ranges.

\begin{figure}[!htb]
\centering\small
\textbf{Grid navigation}\par\smallskip
\begin{minipage}[t]{0.25\linewidth}
\textbf{Component 1 input}
{\scriptsize
\begin{verbatim}
Q:
#######
#S....#
##.#..#
#..##.#
#....G#
#.#.###
#######
A:
\end{verbatim}}
\end{minipage}\hfill
\begin{minipage}[t]{0.25\linewidth}
\textbf{Component 2 input}
{\scriptsize
\begin{verbatim}
Q:
#######
#.....#
#.....#
#....G#
##...##
#S.##.#
#######
A:
\end{verbatim}}
\end{minipage}\hfill
\begin{minipage}[t]{0.4\linewidth}
\textbf{Parent input}
{\scriptsize
\begin{verbatim}
Q:
###############
#S....###.....#
##.#..###.....#
#..##.###....G#
#.......##...##
#.#.###....##.#
###############
A:
\end{verbatim}}
\end{minipage}
\par\smallskip
\begin{tabular}{@{}ll@{}}
Component 1 output: & \texttt{RRRDRDD}\\
Component 2 output: & \texttt{RUURRR}\\
Composed target: & \texttt{RRRDRDD}\,\texttt{RRDRR}\,\texttt{RUURRR} (corridor actions in the middle)
\end{tabular}
\par\medskip\hrule\medskip \textbf{Addition}\par\smallskip
\begin{minipage}[t]{0.3\linewidth}
\textbf{Component 1 input}
{\scriptsize
\begin{verbatim}
Q: 6133 + 8894 = ?
A:
\end{verbatim}}
\end{minipage}\hfill
\begin{minipage}[t]{0.3\linewidth}
\textbf{Component 2 input}
{\scriptsize
\begin{verbatim}
Q: 34355 + 50024 = ?
A:
\end{verbatim}}
\end{minipage}\hfill
\begin{minipage}[t]{0.36\linewidth}
\textbf{Parent input}
{\scriptsize
\begin{verbatim}
Q: 613334355 + 889450024 = ?
A:
\end{verbatim}}
\end{minipage}
\par\smallskip Component outputs \texttt{15027} and \texttt{84379}; composed target \texttt{1502784379}.
\par\medskip\hrule\medskip \textbf{Max run length}\par\smallskip
\begin{minipage}[t]{0.3\linewidth}
\textbf{Component 1 input}
{\scriptsize
\begin{verbatim}
Q: runlen(493036) = ?
A:
\end{verbatim}}
\end{minipage}\hfill
\begin{minipage}[t]{0.3\linewidth}
\textbf{Component 2 input}
{\scriptsize
\begin{verbatim}
Q: runlen(887476) = ?
A:
\end{verbatim}}
\end{minipage}\hfill
\begin{minipage}[t]{0.36\linewidth}
\textbf{Parent input}
{\scriptsize
\begin{verbatim}
Q: runlen(493036887476) = ?
A:
\end{verbatim}}
\end{minipage}
\par\smallskip Composed target \texttt{8|2}, accepted because the boundary symbols \texttt{6} and \texttt{8} differ.
\caption{Saved round-1 labels for the algorithmic tasks, with the prompts the model sees. The composed targets and all inputs are transcribed from the saved label files. Grid child routes are recovered by splitting the saved label at the corridor actions; addition component outputs are the model's saved labels for the same component prompts; max run length component outputs were not saved. All three labels are correct.}
\label{fig:algorithmic_composition_example}
\end{figure}

\subsubsection{Program generation}
\label{appsubsec:bash_setup}
\paragraph{Task.} The input is an instruction listing $k$ steps, each a natural-language description of one text-processing command, and the target is a one-line Bash pipeline that reads standard input and applies the steps in order. A program is correct if its output matches that of the reference pipeline on all $24$ test inputs of its parent.

\paragraph{Composition.} A component is one step, whose answer is one command, and the label joins the component answers with pipes,
\[
A(\hat y_1,\ldots,\hat y_k)=\hat y_1\,\texttt{|}\,\hat y_2\,\texttt{|}\,\cdots\,\texttt{|}\,\hat y_k .
\]
A pipe passes each command's output to the next command's input, so the label is correct whenever every component command is correct (\Cref{fig:bash_composition_example} shows the full prompts).

\paragraph{Filter.} Structural only: the label must parse as a pipeline.

\paragraph{Data.} Commands come from TLDR\footnote{\url{https://github.com/tldr-pages/tldr}} (common and Linux pages) and NL2Bash~\citep{lin_2018_nl2bash}, restricted to $21$ standard text utilities. Of $143$ extracted single-command examples, we keep the $119$ that run deterministically on $24$ probe inputs, produce at least $4$ distinct outputs, and do not act as the identity ($71$ from TLDR, $48$ from NL2Bash). Within each utility, a hash ranking holds out $30$ commands from training compositions, so each held-out command has a seen command of the same utility; all $119$ are still trained and rehearsed as single commands. Pipelines are split into training, validation, and test partitions ($70$, $15$, and $15\%$) by a hash of their unordered command set, so no command set appears in two partitions. Training rounds use $2{,}447$, $4{,}096$, and $4{,}096$ parents of length two, three, and four; the two-command pool has only $2{,}447$ pipelines. The $24$ inputs of a parent come from six seeded generators (logs, CSV, words, key--value pairs, numbers, and tab-separated text) and are shared by all parents of a test set and length. Every checkpoint is evaluated on the $119$ single commands and on three pipeline test sets, split equally between two and three commands: $800$ pipelines of seen commands, $800$ with one held-out command, and $600$ with only held-out commands. Longer pipelines use a separate test set of seen commands, with all $568$ eligible two-command pipelines of the test partition and $1{,}000$ pipelines at each length from three to six. No test pipeline equals a training composition, although $83$--$99\%$ of the three- to six-command test pipelines contain a two-command training composition as a contiguous part.

\begin{figure}[!htb]
\centering
\small \setlength{\tabcolsep}{0pt}
\begin{tabular}{@{}>{\raggedright\arraybackslash}p{\linewidth}@{}}
\toprule
\textbf{Shared instruction block} \\
Use only standard text utilities (such as sort, uniq, grep, cut, tr, head, tail, sed, awk, wc, rev, tac, nl, paste, fold, fmt, cat, column). The text arrives on standard input; do not name any file. Reply with the command on one line and nothing else. \\
\midrule
\textbf{Component 1: model input} \\
Write one Bash command that reads standard input and does the following.\par\smallskip
Task: Output the first 5 lines of a file \\[3pt]
\textbf{Selected output:} \texttt{head -5} \\
\midrule
\textbf{Component 2: model input} \\
Write one Bash command that reads standard input and does the following.\par\smallskip
Task: Sort a file in descending order \\[3pt]
\textbf{Selected output:} \texttt{sort -r} \\
\midrule
\textbf{Parent: model input} \\
Write one Bash pipeline that reads standard input and applies the following steps in order, each step on the output of the previous one, joined with \texttt{|}.\par\smallskip
Step 1: Output the first 5 lines of a file\newline Step 2: Sort a file in descending order \\[3pt]
\textbf{Composed target:} \texttt{head -5 | sort -r} \\
\bottomrule
\end{tabular}
\caption{Saved Bash composition example \texttt{label2\_00059} from round 1. The shared instruction block shown above is inserted after the opening sentence in each prompt; the remaining prompt text and selected outputs are transcribed unchanged. Execution inputs are passed to the program at evaluation and are not part of the prompt. Line wrapping is for display.}
\label{fig:bash_composition_example}
\end{figure}

\subsubsection{Multi-hop reasoning}
\label{appsubsec:musique}
\paragraph{Task.} The input is a MuSiQue question $q$ with its released paragraphs $P$, and the target is a short answer, scored by exact match under the official answer aliases. At evaluation, the model answers each question directly from its paragraphs, without a decomposition.

\paragraph{Composition.} The released decomposition splits $q$ into sub-questions $(q_1,\ldots,q_k)$, $k\in\{2,3,4\}$, where $q_j$ may refer to an earlier answer through a placeholder \texttt{\#}$i$. The model answers the sub-questions in order, seeing its earlier answers and with each placeholder filled by the corresponding answer. For the last step, it answers the original question, with the filled last sub-question given as the final reasoning step (\Cref{fig:musique_composition_example}):
\[
\hat a_j=h\bigl(q_j[\texttt{\#}i\mapsto\hat a_i],\,P,\,\hat a_{<j}\bigr)\ \ (j<k), \qquad A(x)=\hat a_k=h\bigl(q,\,P,\,\hat a_{<k},\,q_k[\texttt{\#}i\mapsto\hat a_i]\bigr),
\]
where $h$ is the current model. Later queries therefore depend on the model's earlier answers, and the rule is valid when the released decomposition leads to the released answer.

\paragraph{Filter.} Every answer $\hat a_j$ must be non-empty, short, free of unresolved placeholders, and present in $P$ or $q$ (\Cref{appsubsec:training_setup}).

\paragraph{Data.} Parents come from the $19{,}938$ answerable MuSiQue training questions, ordered by a fixed hash. We take the first $512$ eligible questions of each hop count, excluding the parents of the seed questions and $22$ questions that match a development question or sub-question; $14{,}019$, $4{,}269$, and $1{,}144$ questions were eligible. The three-hop parents contain $437$ \texttt{3hop1} and $75$ \texttt{3hop2} questions, and the four-hop parents $274$ \texttt{4hop1}, $63$ \texttt{4hop2}, and $175$ \texttt{4hop3} questions. The seed set contains $1{,}024$ single-hop sub-questions with gold-filled references and gold answers, drawn from $484$ training questions. We evaluate on all $2{,}417$ answerable development questions ($1{,}252$ two-hop, $760$ three-hop, and $405$ four-hop); the single-hop row uses $512$ deduplicated development steps.

\begin{figure}[!htb]
\centering\small
\begin{minipage}{\linewidth}
\textbf{Component 1 input} (released first sub-question)
{\scriptsize
\begin{verbatim}
Answer the question using the passages. Reply with the answer only: a name
or short phrase.

[1] Henry Wolf (engraver): Henry Wolf (1852-1916) was a French-born wood
engraver who lived and worked in the United States during his most
influential work period and until his death.

[2] Jeanette Wohl: Jeanette Wohl (October 16, 1783, Frankfurt am Main -
November 27, 1861, Paris) was a longtime friend and correspondent of ...

[... paragraphs 3-20 ...]

Question: Michael Schwerner >> place of death
Answer:
\end{verbatim}}
\textbf{Component 1 output:} \texttt{Mississippi}\par\medskip\hrule\medskip
\textbf{Component 2 input} (final step)
{\scriptsize
\begin{verbatim}
Answer the question using the passages. Reply with the answer only: a name
or short phrase.

[1] Henry Wolf (engraver): ...
[... paragraphs 2-20 ...]

Earlier model predictions (may be incorrect):
#1: Michael Schwerner >> place of death -> Mississippi
Final reasoning step: when did Mississippi become a right to work state

Question: When did the state where Michael Schwerner died become a right to
work state?
Answer:
\end{verbatim}}
\textbf{Component 2 output:} \texttt{1954}\par\medskip\hrule\medskip
\textbf{Parent input}
{\scriptsize
\begin{verbatim}
Answer the question using the passages. Reply with the answer only: a name
or short phrase.

[1] Henry Wolf (engraver): ...
[... paragraphs 2-20 ...]

Question: When did the state where Michael Schwerner died become a right to
work state?
Answer:
\end{verbatim}}
\textbf{Composed target:} \texttt{1954}
\end{minipage}
\caption{Saved multi-hop reasoning label \texttt{2hop\_\_333835\_77060} from the 2B model in round 1. The prompts are rendered from the prompt template with the saved sub-questions and answers; the $20$ paragraphs ($10{,}973$ characters) are truncated for display. The parent input is also the prompt of direct self-improvement. The label is correct.}
\label{fig:musique_composition_example}
\end{figure}

\subsubsection{String transformations}
\paragraph{Task.} Following the controlled setting of \citet{yuan_2025_rl_compositionality}, the input is a source string $x_0$ and an ordered chain of functions $(f_1,\ldots,f_D)$ from a library of $21$ string functions, and the target is the trace $x_i=f_i(x_{i-1})$ for $i=1,\ldots,D$, ending with the final answer $x_D$. The library removes operators of \citet{yuan_2025_rl_compositionality} that duplicate another operator on these inputs or leave most inputs unchanged. The standard configurations use all $21$ functions with $D=4$ and $D=5$; the length-preserving configuration uses the $11$ functions that keep string length unchanged, with $D=4$. A parent is kept only if every intermediate value is nonempty and at most $96$ characters and every function changes the final answer.

\paragraph{Composition.} A component applies one function to the model's previous answer, and the label concatenates the step answers,
\[
\hat x_i=h(f_i,\hat x_{i-1}),\qquad A(x)=(\hat x_1,\ldots,\hat x_D).
\]
As in multi-hop reasoning, each query takes the model's own earlier answer, and the rule is valid on every parent (\Cref{fig:strings_composition_example}). A controller supplies the parent question, the current function, and the model's earlier values at each step; it never executes a function or supplies a correct intermediate value, and the same controller is used for both methods in training and evaluation.

\paragraph{Filter.} Structural only: each value must parse and have at most $96$ characters.

\paragraph{Data.} The initial model is trained on $200{,}000$ chains of one function and $200{,}000$ chains of two functions, covering all $21$ functions and all $441$ ordered pairs, with lowercase sources of $4$--$8$ characters. Each parent has one fresh lowercase source of $4$--$10$ characters and functions drawn with replacement. Parents are assigned to $2{,}048$ training, $128$ development, and $512$ test parents by a hash of their multiset of functions, so every reordering of the same functions falls in one split; multisets used in earlier screening are excluded.

\begin{figure}[!htb]
\centering\small
\begin{tabular}{@{}lll@{}}
\toprule
 & Model input & Saved output \\
\midrule
Component 1 & \texttt{Q: func\_2(bnqi) = ?} & \texttt{func\_2=binq =\textgreater{} binq} \\
Component 2 & \texttt{Q: func\_15(binq) = ?} & \texttt{func\_15=elqt =\textgreater{} elqt} \\
Component 3 & \texttt{Q: func\_15(elqt) = ?} & \texttt{func\_15=hotw =\textgreater{} hotw} \\
Component 4 & \texttt{Q: func\_4(hotw) = ?} & \texttt{func\_4=hOtW =\textgreater{} hOtW} \\
Component 5 & \texttt{Q: func\_13(hOtW) = ?} & \texttt{func\_13=WhOt =\textgreater{} WhOt} \\
\bottomrule
\end{tabular}
\par\smallskip
\begin{minipage}{\linewidth}
\textbf{Parent input}
{\scriptsize
\begin{verbatim}
Q: func_13(func_4(func_15(func_15(func_2(bnqi))))) = ?
A:
\end{verbatim}}
\textbf{Composed target}
{\scriptsize
\begin{verbatim}
func_2=binq | func_15=elqt | func_15=hotw | func_4=hOtW | func_13=WhOt => WhOt
\end{verbatim}}
\end{minipage}
\caption{Saved compositional warmup label for strings of depth five. Each input is written as \texttt{Q: \dots\ = ?} followed by a new line and \texttt{A:}; each component input applies one function to the model's previous output. The parent input is also the prompt of direct RL. All text is transcribed from the saved files, and the label is correct.}
\label{fig:strings_composition_example}
\end{figure}

\subsubsection{Typed Python}
\paragraph{Task.} The input is the natural-language specification of three type-compatible unary functions $(g_1,g_2,g_3)$, and the target is the body of \texttt{solve(x)} that applies them in order. A program is correct if it passes all $24$ test inputs of its parent.

\paragraph{Composition.} A component asks for a program $\hat p_i$ for one specification, and the label places the component programs in a fixed wrapper (\Cref{fig:python_composition_example}),
\[
A(\hat p_1,\hat p_2,\hat p_3)=\bigl[\,\texttt{solve(x)}:\ \texttt{return}\ \hat p_3(\hat p_2(\hat p_1(\texttt{x})))\,\bigr],
\]
so the label is correct whenever every component program is correct on its inputs.

\paragraph{Filter.} Structural only: the label must parse as one Python function.

\paragraph{Data.} Functions come from MBPP+ and HumanEval+ \citep{csi_mbpp_2021,chen_2021_humaneval,csi_evalplus_2023}: $189$ unary functions over integers, strings, and lists with at least $48$ extended test inputs and at least $4$ distinct outputs. The initial model is Qwen3.5-2B after $128$ fine-tuning updates on its own single-function programs that pass $12$ acceptance inputs, and we use the $106$ functions it then solves reliably. Parents are chains with matching input and output types and no repeated function, assigned to $2{,}048$ training, $128$ development, and $512$ test parents by a hash of their function multiset. The $24$ inputs of a parent are extended test inputs of its first function on which the whole chain runs with the expected types; the first $12$ form the RL reward, and all $24$ are used for evaluation.

\begin{figure}[!htb]
\centering\small
\begin{minipage}{\linewidth}
\textbf{Shared instruction block}
{\fontsize{5.8}{6.8}\selectfont
\begin{verbatim}
Complete the body of the Python function solve(x). The response already
starts with `def solve(x):` and indentation; continue with its body without
repeating the header. You may import standard-library modules and define
helper functions inside solve. Apply the following operations in order,
passing each output as the next input. Return the final value. Inputs satisfy
the listed preconditions at every step. Output only the Python continuation,
with correct indentation and no Markdown fences.
\end{verbatim}}
\end{minipage}\par\smallskip\hrule\smallskip
\begin{minipage}[t]{0.45\linewidth}
\textbf{Component 1 input}
{\fontsize{5.8}{6.8}\selectfont
\begin{verbatim}
Step 1:
"""
Write a function to remove lowercase substrings
from a given string.
assert remove_lowercase("PYTHon")==('PYTH')
"""

Input preconditions:
assert isinstance(str1, str), "invalid inputs"
\end{verbatim}}
\textbf{Component 1 output}
{\fontsize{5.8}{6.8}\selectfont
\begin{verbatim}
def solve(x):
    if not isinstance(x, str):
        raise ValueError('Input must be a string')
    return ''.join((c for c in x if c.isupper()))
\end{verbatim}}
\textbf{Component 2 input}
{\fontsize{5.8}{6.8}\selectfont
\begin{verbatim}
Step 1:
"""
Write a python function to convert the given
string to lower case.
assert is_lower("InValid") == "invalid"
"""
\end{verbatim}}
\textbf{Component 2 output}
{\fontsize{5.8}{6.8}\selectfont
\begin{verbatim}
def solve(x):
    return x.lower()
\end{verbatim}}
\textbf{Component 3 input}
{\fontsize{5.8}{6.8}\selectfont
\begin{verbatim}
Step 1:
"""
Write a python function to split a string into
characters.
assert split('python') == ['p','y','t','h','o','n']
"""
\end{verbatim}}
\textbf{Component 3 output}
{\fontsize{5.8}{6.8}\selectfont
\begin{verbatim}
def solve(x):
    if not isinstance(x, str):
        raise ValueError('invalid inputs')
    return list(x)
\end{verbatim}}
\end{minipage}\hfill
\begin{minipage}[t]{0.53\linewidth}
\textbf{Parent input}\par
The instruction block followed by the three step blocks, numbered Steps 1--3.\par\medskip
\textbf{Composed target}
{\fontsize{5.8}{6.8}\selectfont
\begin{verbatim}
def solve(x):
    def _component_0(_value):

        def solve(x):
            if not isinstance(x, str):
                raise ValueError('Input must be a string')
            return ''.join((c for c in x if c.isupper()))
        return solve(_value)
    def _component_1(_value):

        def solve(x):
            return x.lower()
        return solve(_value)
    def _component_2(_value):

        def solve(x):
            if not isinstance(x, str):
                raise ValueError('invalid inputs')
            return list(x)
        return solve(_value)
    value = x
    value = _component_0(value)
    value = _component_1(value)
    value = _component_2(value)
    return value
\end{verbatim}}
\end{minipage}
\caption{Saved compositional warmup label for typed Python (MBPP+ functions 478, 477, and 565). The response of every prompt begins with \texttt{def solve(x):}, which the model continues; the component preconditions of Steps 2 and 3 are omitted for space. All text is transcribed from the saved files, and the composed program passes all $24$ test inputs of its parent.}
\label{fig:python_composition_example}
\end{figure}

\clearpage
\subsection{Training Setup}
\label{appsubsec:training_setup}
This subsection gives the training setup of every experiment. In all of them, every round runs to completion and no checkpoint is selected on validation data.

\paragraph{Algorithmic tasks (\Cref{sec:setup}).} All three tasks use a $14.2$M-parameter Transformer with six layers, hidden size $384$, six attention heads, and no positional encoding, trained with AdamW ($\beta_1=0.9$, $\beta_2=0.99$, $\epsilon=10^{-12}$) in bfloat16. Labels are generated with greedy decoding. \Cref{tab:training_setup} lists the seed training, round schedule, composed and replay data, and filter of each task. For the component-accuracy study of grid navigation (\Cref{fig:grid_seed_accuracy}),
\label{appsubsec:grid_seed_accuracy}%
we train three initial models with different seeds (batch size $128$) and run the grid schedule of \Cref{tab:training_setup} from their checkpoints after $400$, $1{,}500$, $2{,}500$, $3{,}000$, $6{,}000$, $8{,}000$, and $30{,}000$ updates.

\begin{table}[!htb]
\centering\footnotesize
\setlength{\tabcolsep}{3pt} \renewcommand{\arraystretch}{1.15}
\caption{Training setup for the algorithmic tasks.}
\label{tab:training_setup}
\begin{tabular}{@{}>{\raggedright\arraybackslash}p{0.16\textwidth}
                >{\raggedright\arraybackslash}p{0.25\textwidth}
                >{\raggedright\arraybackslash}p{0.25\textwidth}
                >{\raggedright\arraybackslash}p{0.25\textwidth}@{}}
\toprule
 & Grid navigation & Addition & Max run length \\
\midrule
Model & \multicolumn{3}{l}{$14.2$M-parameter character-level Transformer} \\
Learning rate & $5\times10^{-4}$ & $5\times10^{-4}$ & $5\times10^{-4}$ \\
Weight decay & $0.1$ & $0.1$ & $0.1$ \\
Batch size & $256$ & $256$ & $256$ \\
Initial training & $200{,}000$ one-room grids, $30{,}000$ updates & 3--7 digits, $50{,}000$ per length, $10{,}000$ updates & Lengths 6--10, $50{,}000$ per length, $15{,}000$ updates \\
Rounds & 12 & 8 & 8 \\
Range & 1 to 13 rooms, $+1$ per round & 7 to 23 digits, $+2$ per round & Length 10 to 83: $20$ after round 1, then $+9$ per round \\
Composed examples & $15{,}000$ per room count up to the maximum, redrawn each round & $10{,}000$ per length up to the maximum, redrawn each round & $2{,}000$ per length up to the maximum, redrawn each round \\
Replay at initial levels & $5{,}000$ & $25{,}000$& $250{,}000$ \\
Filter & None & No boundary carry & Boundary symbols differ \\
Training per round & $9{,}000$ updates & $3{,}000$ updates & $3{,}000$ updates \\
Test set & $200$ per room count, $1$--$13$ & $100$ per length, $3$--$23$ & $100$ per length, $6$--$83$ \\
\bottomrule
\end{tabular}
\end{table}

\paragraph{Program generation and multi-hop reasoning (\Cref{sec:programming_reasoning}).} Both experiments fine-tune Qwen3.5 models with LoRA (\Cref{tab:realworld_hparams}). Initial training uses components with reference answers: single commands for Bash and single-hop questions for MuSiQue. Each round labels its parents with the model from the preceding round, continues training from that model, and mixes the new labels with rehearsal of the initial examples. Labels are never executed or checked against reference answers; only the filters of \Cref{appsubsec:comp_rule} are applied.

\begin{table}[!htb]
\centering\small
\caption{Hyperparameters for the experiments in \Cref{sec:programming_reasoning}.}
\label{tab:realworld_hparams}
\label{tab:bash_hparams}\label{tab:musique_hparams}
\begin{tabular}{@{}l>{\raggedright\arraybackslash}p{0.36\textwidth}>{\raggedright\arraybackslash}p{0.36\textwidth}@{}}
\toprule
 & Program generation & Multi-hop reasoning \\
\midrule
Model & Qwen3.5-2B; LoRA rank $16$, $\alpha=32$ & Qwen3.5-2B, 4B, 9B; LoRA rank $16$ on attention and feed-forward projections \\
Initial training & All $119$ commands, $96$ updates, learning rate $10^{-4}$ & $1{,}024$ single-hop questions, $100$ updates, batch size $16$, learning rate $2\times10^{-4}$ \\
Rounds & $3$ (pipelines of $2$, $3$, $4$ commands) & $3$ ($2$-, $3$-, $4$-hop questions) \\
Parents per round & $2{,}447$, $4{,}096$, $4{,}096$ & $512$ \\
Training per round & $192$ AdamW updates, learning rate $10^{-4}$, batch size $16$, gradient clipping $1.0$ & $64$ AdamW updates, learning rate $10^{-4}$, batch size $8$ \\
Rehearsal & $25\%$ of each batch & $2$ of $8$ examples per batch \\
Seeds & $4$ & $3$ \\
\bottomrule
\end{tabular}
\end{table}

\paragraph{RL warm start (\Cref{sec:rl_warm_start}).}
\label{appsubsec:rl_setup}%
The compositional warmup labels the first $512$ training parents, in a fixed hash order, with the initial model and fine-tunes on them (\Cref{appsubsec:comp_rule}); RL samples prompts from all $2{,}048$ training parents. Both methods then run GRPO \citep{shao_2024_deepseekmath} with the configuration released by \citet{yuan_2025_rl_compositionality} (\Cref{tab:grpo}). The reward is $1$ for a correct final answer and $0$ otherwise; for Python, it uses $12$ of the $24$ inputs of each parent, and evaluation uses all $24$ with greedy decoding. Generated tokens include warmup label generation and every sampled RL token, including groups discarded for uniform reward, and the learning curves also charge the target tokens of the warmup fine-tuning; prompt tokens and evaluation are excluded. On strings, the warmup-then-RL runs generate $40$M tokens and direct RL $80$M, with checkpoints every $10$M tokens up to $40$M and at $80$M. On Python, they generate $524{,}288$ and $1{,}048{,}576$ tokens, with checkpoints every $262{,}144$ tokens; continuations restore the optimizer and random-number state without new labels, so label costs are charged once. The reported checkpoints are fixed in advance, and three seeds share the initial model and data.

\begin{table}[t]
\centering\small
\caption{Hyperparameters for the RL experiments in \Cref{sec:rl_warm_start}.}
\label{tab:grpo}
\begin{tabular}{@{}l>{\raggedright\arraybackslash}p{0.27\textwidth}>{\raggedright\arraybackslash}p{0.27\textwidth}@{}}
\toprule
 & Strings & Typed Python \\
\midrule
\multicolumn{3}{@{}l}{\emph{Initial model and compositional warmup}} \\
Initial model & $14.2$M character Transformer & Qwen3.5-2B, LoRA rank $16$, $\alpha=32$ \\
Initial training & Chains of $1$--$2$ functions & $128$ updates on single functions \\
Warmup SFT & $24$ updates, learning rate $10^{-4}$ & $24$ updates, learning rate $10^{-5}$ \\
Target tokens per update & $8{,}192$ & $4{,}096$ \\
\midrule
\multicolumn{3}{@{}l}{\emph{GRPO}} \\
Optimizer, learning rate & \multicolumn{2}{l}{AdamW, $10^{-5}$} \\
Samples per group & $16$ & $8$ \\
Groups with varied rewards per update & $16$ & $4$ \\
Sampling temperature & \multicolumn{2}{l}{$1.0$} \\
Maximum generation length & $96$ characters per step & $4{,}096$ tokens \\
Clip ratio, KL coefficient & \multicolumn{2}{l}{$0.2$, $0$} \\
\bottomrule
\end{tabular}
\end{table}

\section{Theoretical Analysis}\label{app:theory_proofs}
When can predictions on simpler problems provide useful supervision for harder ones? We examine three requirements: reliable component predictions, valid composition, and sufficient supervision coverage. We first analyze how component and composition errors affect label reliability, then characterize the limitations of supervision coverage. Finally, we give a sufficient condition under which training on composed labels improves accuracy on accepted examples.

\subsection{Setup: task levels and composed supervision}
\label{appsubsec:level_setup}

\paragraph{Task levels.}
Let $f^\star:\mathcal X\to\mathcal Y$ be the target function. We use finite input, answer, and probability spaces with exact-match loss. A task-specific level $\ell:\mathcal X\to\mathbb N$ indexes problem scale. Write
\begin{equation}\label{eq:csi_level_sets}
\mathcal X_k=\{x:\ell(x)=k\},
\qquad
\mathcal X_{<k}=\{x:\ell(x)<k\}.
\end{equation}
Levels describe problem scale, without assuming that lower-level problems are solved or that accuracy decreases with level. For example, a shorter addition problem requiring more carry propagation may be harder than a longer one. Initial training may cover only part of a level, and subsequent rounds may revisit earlier levels while introducing new ones.

Fix a level $k$, the current model $\hat f$, and an available source pool $S\subseteq\mathcal X_{<k}$. We consider decompositions whose component queries remain in $S$, both with model-predicted and correct intermediate answers. For decompositions with fixed queries, this requires $\ell(x_i)<k=\ell(x)$ for each component. We use this level ordering to analyze supervision coverage.

\paragraph{Candidate and training examples.}
Let $D_k$ be the distribution of candidate problems at level $k$. These problems can be generated by composing simpler problems or selected from existing problems with supplied decompositions. For each candidate with an available decomposition, the current model constructs a composed pseudo-label, and a filter $G\in\{0,1\}$ determines whether the example is retained for training. Let $\Pr_k$ denote probability over this sampling and labeling process, conditional on the current model and training history, and let $q_k:=\Pr_k(G=1)>0$ be the probability that a sampled candidate is retained. For example, in addition, $D_k$ samples problems at level $k$, and the filter rejects compositions with carries across block boundaries. Here, $q_k$ measures the fraction of candidate compositions that pass this check.

Importantly, $D_k$ need not cover all problems in $\mathcal X_k$. Our analysis therefore concerns the candidate distribution and, when conditioned on $G=1$, the examples retained for training. The resulting guarantees do not necessarily extend to all problems at level $k$ or to other evaluation distributions.

\paragraph{Comparison with correct components.}
Consider a candidate problem $X$ with a decomposition $\pi$. The current model predicts answers to its subproblems, which are combined by the composition rule $A$ to produce a pseudo-label $\widetilde Y$. To distinguish component errors from composition errors, let $\widetilde Y^\circ$ denote the composed label obtained using correct component answers instead of the model's predictions. This label may still be incorrect if the composition rule is invalid. For sequential tasks, such as string transformations, correcting intermediate answers also changes the inputs to subsequent steps. We compare the two executions using the same initial state and non-component random choices, assuming a finite number of component calls. For fixed queries and deterministic composition, $\widetilde Y^\circ$ is given by
\[
\widetilde Y^\circ =A\bigl(\pi,f^\star(X_1),\ldots,f^\star(X_m)\bigr).
\]

When all component predictions are correct, $\widetilde Y=\widetilde Y^\circ$. Let $E_{\mathrm{comp}}$ denote the event that at least one component prediction is incorrect on the query actually issued. We then define three error rates:
\begin{equation}\label{eq:accepted_error_definitions}
\begin{aligned}
\rho_k&=\Pr_k[\widetilde Y\ne f^\star(X)\mid G=1],\\
\epsilon_k&=\Pr_k[E_{\mathrm{comp}}\mid G=1],\\
\xi_k&=\Pr_k[\widetilde Y^\circ\ne f^\star(X)\mid G=1].
\end{aligned}
\end{equation}
Here $\rho_k$ measures the overall composed-label error, $\epsilon_k$ the probability of any component error, and $\xi_k$ the error remaining even with correct components. All three are evaluated on examples accepted by the original filter $G$, so they refer to the same training population. When composition uses a learned aggregator, $\xi_k$ also includes its errors with correct component answers.

\subsection{Reliability and Failure Modes of Composed Supervision}
\label{appsubsec:reliability_failure_modes}

We first relate composed-label error to component and composition errors. We then examine the three failure modes: unreliable component predictions, invalid composition rules, and limited supervision coverage.

Correcting the component answers leaves the composed label unchanged whenever those answers were already correct. This observation gives a reliability bound that allows correlated component mistakes and prediction-dependent filtering.

\begin{proposition}[Reliability at a task level]
\label[proposition]{prop:k_component_noise}
The composed-label error satisfies
\begin{equation}\label{eq:csi_reliability_comparison}
|\rho_k-\xi_k|\leq\epsilon_k,
\end{equation}
and hence
\begin{equation}\label{eq:csi_reliability_bound}
\rho_k\leq\min\{1,\epsilon_k+\xi_k\}.
\end{equation}
If each execution uses at most $m$ component calls, let $\epsilon_{k,i}$ be the probability, conditional on $G=1$, that call $i$ occurs and its answer is incorrect. Then
\[
\epsilon_k\leq\sum_{i=1}^{m}\epsilon_{k,i}.
\]
No independence between component mistakes is required.
\end{proposition}

\begin{proof}
The two labels coincide whenever all component predictions are correct. Therefore,
\[
\left| \mathbf{1}\{\widetilde Y\ne f^\star(X)\} - \mathbf{1}\{\widetilde Y^\circ\ne f^\star(X)\} \right| \leq \mathbf{1}\{E_{\mathrm{comp}}\}.
\]
Taking expectations conditional on $G=1$ and applying the triangle inequality gives $|\rho_k-\xi_k|\leq\epsilon_k$. The upper bound on $\rho_k$ follows immediately, and a conditional union bound gives the bound on $\epsilon_k$. When later queries depend on earlier answers, the two executions remain identical in the absence of component errors: each correct answer produces the same subsequent query under the shared composition procedure and random choices.
\end{proof}

\paragraph{When composition improves the supervision.}
To compare composed and direct labels, evaluate both on the same accepted candidates. Let
\[
\rho_k^{\mathrm{direct}} = \Pr_k[\hat f(X)\ne f^\star(X)\mid G=1]
\]
be the current model's direct prediction error on these problems. By \Cref{prop:k_component_noise}, a sufficient condition for composed labels to be more accurate is
\begin{equation}\label{eq:csi_supervision_advantage}
\epsilon_k+\xi_k<\rho_k^{\mathrm{direct}}.
\end{equation}
Thus, composition need not produce perfect labels to improve supervision: its combined component and composition errors need only fall below the model's direct error. The comparison must use the same candidates. Evaluating the two labeling methods on separately filtered populations does not establish this advantage.

We now examine these three failure modes in turn.
\paragraph{Failure mode I: Low component accuracy.}
\label{appsubsec:component_failure}
Composed supervision relies on accurate predictions across all components. Even when individual components are reliable, their errors can accumulate during composition. The extent of this problem depends not only on component accuracy, but also on how their errors overlap across problems. We characterize this relationship without assuming independent errors.

\begin{proposition}[Error propagation in composed supervision]
\label[proposition]{prop:csi_component_obstruction}
For fixed-query deterministic composition, suppose the composition rule is valid for every accepted candidate and that component errors are not masked: whenever at least one component prediction is incorrect, the composed answer differs from that obtained with correct components. Then
\begin{equation}\label{eq:csi_component_error_identity}
\rho_k=\epsilon_k,
\end{equation}
and
\begin{equation}\label{eq:csi_component_error_bounds}
\max_{1\leq i\leq m}\epsilon_{k,i}
\leq \rho_k
\leq \min\!\left\{1,\sum_{i=1}^{m}\epsilon_{k,i}\right\}.
\end{equation}
\end{proposition}

\begin{proof}
Since the composition rule is valid, correct component predictions produce a correct label. Under the no-masking assumption, any component error makes the composed label incorrect. Thus, $\rho_k=\epsilon_k$, the probability of at least one component error. This probability is at least that of any individual component error and at most their sum. The lower bound is attained when error events are nested, and the upper bound when they are disjoint or jointly cover all accepted examples.
\end{proof}

The main implication is that accurate component predictions are essential for reliable composed supervision. Even with a valid composition rule, a single incorrect component can make the composed label wrong. However, individual component accuracy does not fully determine label reliability: how component errors overlap across problems also matters.

 For example, consider ten components, each with 90\% accuracy. If their errors occur on the same 10\% of problems, composed-label accuracy remains 90\%, attaining the lower error bound. If their errors occur on disjoint subsets, every composed label is incorrect, attaining the upper bound. Thus, even highly accurate components can produce unreliable supervision when their errors accumulate across different problems.

\paragraph{Failure mode II: Composition errors.}
\label{appsubsec:structural_failure}
Even perfect component predictions cannot guarantee correct composed labels when the composition rule is invalid. Recall that $\xi_k$ measures the error of the composition procedure when all component answers are correct. By \Cref{eq:csi_reliability_comparison},
\begin{equation}\label{eq:csi_structural_error_floor}
\rho_k\geq\max\{0,\xi_k-\epsilon_k\}.
\end{equation}
In particular,
\begin{equation}\label{eq:csi_perfect_components}
\epsilon_k=0
\quad\Longrightarrow\quad
\rho_k=\xi_k.
\end{equation}
Thus, improving component accuracy alone cannot eliminate errors introduced by an invalid composition rule. When component errors are rare, composed-label error remains close to $\xi_k$. If $\xi_k-\epsilon_k\geq\rho_k^{\mathrm{direct}}$, composed supervision cannot outperform direct predictions on the same accepted examples.

Unlike Failure mode I, addressing this limitation requires correcting the composition rule or filtering out invalid compositions. A \emph{structurally sound} filter retains only examples for which composition yields the correct answer when all component predictions are correct, ensuring $\xi_k=0$. However, filters based on uncertain model predictions may not provide this guarantee. Moreover, component errors can occasionally cancel errors introduced by an invalid rule, so correcting component predictions does not always improve label accuracy. Thus, filtering offers a way to remove structural errors, but at a cost: excluding invalid compositions may leave important problems without supervision. This leads to our third failure mode.

\paragraph{Failure mode III: Limited supervision coverage.}
\label{appsubsec:coverage_failure}
Accurate composed labels may still provide little supervision for some output patterns required by harder problems. This limitation concerns both the construction of supervision and its effect on learning: composition may rarely supply certain outputs, and ignoring those outputs may incur little penalty during training. We characterize these two aspects below.

Let $C\subseteq\mathcal Y$ be a nonempty set of commonly supervised outputs, and let
\[
U_{k,C}=\{x\in\mathcal X_k:f^\star(x)\notin C\}
\]
contain the level-$k$ problems requiring outputs outside $C$. Write
\[
p_{k,C}=\Pr_k[X\in U_{k,C}\mid G=1]
\]
for the fraction of accepted examples belonging to this group. We allow $p_{k,C}$ to be small but positive, rather than assuming that these problems are entirely absent from supervision.

Consider deterministic composition whose component queries are fixed by the decomposition. Suppose the rule preserves $C$:
\begin{equation}\label{eq:csi_output_closure}
A(\pi,y_1,\ldots,y_m)\in C
\quad\text{whenever }y_1,\ldots,y_m\in C
\text{ and }\pi\text{ is allowed}.
\end{equation}
With correct component answers, producing a label outside $C$ therefore requires at least one component answer outside $C$.

\begin{proposition}[Coverage depends on component answers]
\label[proposition]{prop:csi_output_coverage}
Under the preceding closure condition,
\begin{equation}\label{eq:csi_uncovered_retention}
p_{k,C}
\leq
\xi_k+
\sum_{i=1}^{m}
\Pr_k[f^\star(X_i)\notin C\mid G=1].
\end{equation}
For a structurally sound filter, $\xi_k=0$. The fraction of accepted problems requiring outputs outside $C$ is therefore bounded by how often their components require such outputs.
\end{proposition}

\begin{proof}
If all correct component answers lie in $C$, closure ensures that $\widetilde Y^\circ\in C$. Thus, an accepted problem in $U_{k,C}$ must either have an incorrect composed label even with correct components, or have at least one component whose correct answer lies outside $C$. A union bound conditional on $G=1$ proves the result.
\end{proof}

Thus, composition need not distribute supervision evenly across the output patterns required by harder problems. When the sum in \Cref{eq:csi_uncovered_retention} is small, a structurally sound filter retains few problems requiring outputs outside $C$. If no sampled component supplies such an answer, their coverage is zero. Generating harder inputs alone does not remove this restriction; the construction must also draw on components or rules that can supply the missing outputs.

\medskip
\noindent\emph{Why limited coverage can leave errors unresolved.} A predictor that never produces outputs outside $C$ fails on every problem in $U_{k,C}$. Yet this restriction may cost little when fitting the composed supervision. To make this precise, consider a finite hypothesis class $\mathcal H$ and define
\[
L_k(h)=\Pr_k[h(X)\ne\widetilde Y\mid G=1],
\]
the exact-match error against accepted composed labels. Let
\[
\mathcal H_C = \{h\in\mathcal H:h(x)\in C \text{ for every }x\in\mathcal X_k\}
\]
contain the predictors whose level-$k$ outputs are restricted to $C$. The following result isolates coverage from label errors by assuming that the accepted labels are correct.

\begin{proposition}[The training penalty for ignoring output patterns]
\label[proposition]{prop:csi_absent_output_penalty}
Suppose $\rho_k=0$ and $f^\star\in\mathcal H$. Assume that, for any $h\in\mathcal H$, replacing all of its level-$k$ predictions outside $C$ with a fixed answer in $C$ gives another predictor in $\mathcal H$. Then
\begin{equation}\label{eq:csi_output_restriction_penalty}
\min_{h\in\mathcal H_C}L_k(h)
-
\min_{h\in\mathcal H}L_k(h)
=
p_{k,C}.
\end{equation}
Nevertheless, every $h\in\mathcal H_C$ satisfies
\begin{equation}\label{eq:csi_restricted_output_error}
\Pr_{X\sim D_k}[h(X)\ne f^\star(X)]
\geq D_k(U_{k,C}).
\end{equation}
In particular, if $p_{k,C}=0$, a predictor can minimize $L_k$ while being wrong on every problem in $U_{k,C}$.
\end{proposition}

\begin{proof}
Since the accepted labels are correct and $f^\star\in\mathcal H$, the minimum unrestricted error is zero. Every predictor in $\mathcal H_C$ is wrong whenever the correct answer lies outside $C$, giving $L_k(h)\geq p_{k,C}$.

To attain this bound, replace the level-$k$ outputs of $f^\star$ outside $C$ with a fixed answer in $C$. By assumption, the resulting predictor belongs to $\mathcal H_C$. It is correct outside $U_{k,C}$ and wrong throughout $U_{k,C}$, so its error on accepted examples is exactly $p_{k,C}$. This proves \Cref{eq:csi_output_restriction_penalty}. Every predictor in $\mathcal H_C$ also fails on all candidate problems in $U_{k,C}$, giving \Cref{eq:csi_restricted_output_error}.
\end{proof}

The result explains how high average training accuracy can coexist with systematic failure on underrepresented outputs. If $p_{k,C}$ is small, a predictor that never produces those outputs can be nearly optimal on the accepted distribution, despite having zero accuracy on the corresponding problems. For example, if these problems constitute only $0.1\%$ of accepted examples, ignoring them need cost only $0.1\%$ in average exact-match accuracy. If $p_{k,C}=0$, the objective does not distinguish the restricted predictor from the correct one at all.

Together, the two results connect supervision construction to this learning limitation. Composition can provide accurate labels while giving little or no weight to certain required outputs; fitting those labels then imposes little or no penalty for never producing them. More examples can help when coverage is positive, while increasing the representation of these patterns strengthens their contribution to the objective. Additional examples confined to $C$ cannot resolve the zero-coverage case.

\subsection{From Composed Supervision to Model Improvement}
\label{appsubsec:finite_learning}

The preceding analysis characterizes when composed supervision is reliable and what problems it can cover. We now ask whether training on these labels actually improves the model. We analyze a single training round, deriving conditions under which composed supervision improves prediction accuracy on accepted examples. We then examine how limited supervision coverage constrains this guarantee.

Let $R_k^+(h)$ denote the true-label error on accepted examples, and recall that $L_k(h)$ denotes the corresponding pseudo-label error.
\[
R_k^+(h)=\Pr_k[h(X)\ne f^\star(X)\mid G=1], \qquad L_k(h)=\Pr_k[h(X)\ne\widetilde Y\mid G=1].
\]
For every predictor $h$,
\begin{equation}\label{eq:csi_pseudorisk_comparison}
|R_k^+(h)-L_k(h)|\leq\rho_k,
\end{equation}
and also
\begin{equation}\label{eq:teacher_imitation_lower_bound}
|R_k^+(h)-\rho_k|\leq L_k(h).
\end{equation}
The first inequality follows because true-label and pseudo-label losses coincide when the pseudo-label is correct. The second follows because the student and pseudo-labeler have the same correctness whenever their answers agree. In particular,
\begin{equation}\label{eq:csi_structural_student_lower_bound}
R_k^+(h)\geq\xi_k-\epsilon_k-L_k(h).
\end{equation}
When the composition rule is valid and component errors are not masked, \Cref{eq:csi_component_error_identity} gives
\begin{equation}\label{eq:csi_component_student_lower_bound}
R_k^+(h)\geq\epsilon_k-L_k(h).
\end{equation}
A student that closely matches the composed pseudo-labels also inherits most of their errors. Correcting these errors requires deviating from some pseudo-labels. Fitting a finite training set alone does not establish a small population imitation loss $L_k(h)$.

\begin{samepage}
\begin{theorem}[Learning on supervised examples at a level]
\label{thm:finite_class_noisy_erm}
Fix a nonempty finite class $\mathcal H$ containing $f^\star$ before sampling. Draw $n\geq1$ independent examples from the conditional law of $(X,\widetilde Y)$ given $G=1$. Suppose $\hat f_{\mathrm{new}}\in\mathcal H$ minimizes empirical exact-match pseudo-label error up to an additive $\tau\geq0$. For $\delta\in(0,1)$, with probability at least $1-\delta$,
\begin{equation}\label{eq:csi_level_learning_bound}
R_k^+(\hat f_{\mathrm{new}})
\leq 2\rho_k+
2\sqrt{\frac{\log(2|\mathcal H|/\delta)}{2n}}+\tau.
\end{equation}
\end{theorem}
\end{samepage}
\begin{proof}
Let $u_n=\sqrt{\log(2|\mathcal H|/\delta)/(2n)}$. Hoeffding's inequality \citep{hoeffding_1963_probability} and a union bound give uniform deviation at most $u_n$ between empirical and population pseudo-label risks. On this event, approximate minimization and $f^\star\in\mathcal H$ imply $L_k(\hat f_{\mathrm{new}})\leq L_k(f^\star)+2u_n+\tau =\rho_k+2u_n+\tau$. \Cref{eq:csi_pseudorisk_comparison} proves the result.
\end{proof}

The theorem shows that reliable composed supervision can translate into improved predictions, provided that sampling and optimization errors are sufficiently small. More training examples and better optimization tighten the guarantee, but the bound remains dependent on pseudo-label accuracy.

\paragraph{A sufficient condition on the supervised population.}
Combining the theorem with \Cref{prop:k_component_noise} certifies improvement over the current model on the same accepted parent population whenever
\begin{equation}\label{eq:csi_sufficient_improvement}
2(\epsilon_k+\xi_k)
+2\sqrt{\frac{\log(2|\mathcal H|/\delta)}{2n}}+\tau
<R_k^+(\hat f).
\end{equation}
The condition separates label error, sampling error, and optimization error. It is sufficient for improvement on accepted parents, without pointwise noise assumptions or independence among component mistakes. It does not guarantee improvement on excluded problems.

\paragraph{Evaluation across the level.}
To evaluate a predictor on the pre-filter level-$k$ population, write $R_k(h)=\Pr_{X\sim D_k}[h(X)\ne f^\star(X)]$. When $0<q_k<1$, the law of total probability gives
\begin{equation}\label{eq:csi_level_risk_split}
R_k(h)=q_k R_k^+(h)
+(1-q_k)\Pr_k[h(X)\ne f^\star(X)\mid G=0].
\end{equation}
For $q_k=1$, $R_k(h)=R_k^+(h)$. The theorem controls error on accepted examples; error on rejected examples remains unconstrained. Bounding the latter by one gives the conservative full-population bound
\[
R_k(\hat f_{\mathrm{new}}) \leq q_k\min\{1,2\rho_k+2u_n+\tau\}+(1-q_k),
\]
where $u_n$ is defined in the proof above. Thus high retention limits the mass left uncontrolled under $D_k$. It says nothing about problems that $D_k$ omits or about a different evaluation law.

\clearpage
\section{Additional Experimental Results}\label{appsec:additional_experiments}

\subsection{Direct Self-Improvement and Filtering}
\label{appsubsec:algorithmic_baselines}
\Cref{fig:self_improvement_panels} shows compositional self-improvement alone. This subsection compares it with direct self-improvement on the three algorithmic tasks and, for addition and max run length, with composition without the filter. It then shows the Bash and RL warm-start comparisons of \Cref{subsec:extension_sft,sec:rl_warm_start} alike.

\paragraph{Setup.}
For each task, all runs share the initial model, parent inputs, round schedule, and replay at the initial difficulty levels specified in \Cref{tab:training_setup}; the runs differ only in how new labels are produced. Direct self-improvement labels each parent with the current model's greedy prediction on the full parent. Unfiltered composition keeps every composed label. Filtered composition applies the filters of \Cref{tab:controlled_composition_examples}, which reject addition pairs with a carry across a block boundary and max run length pairs with equal boundary symbols. Grid navigation has no filter.

\begin{figure}[!htb]
\centering
\includegraphics[width=\textwidth]{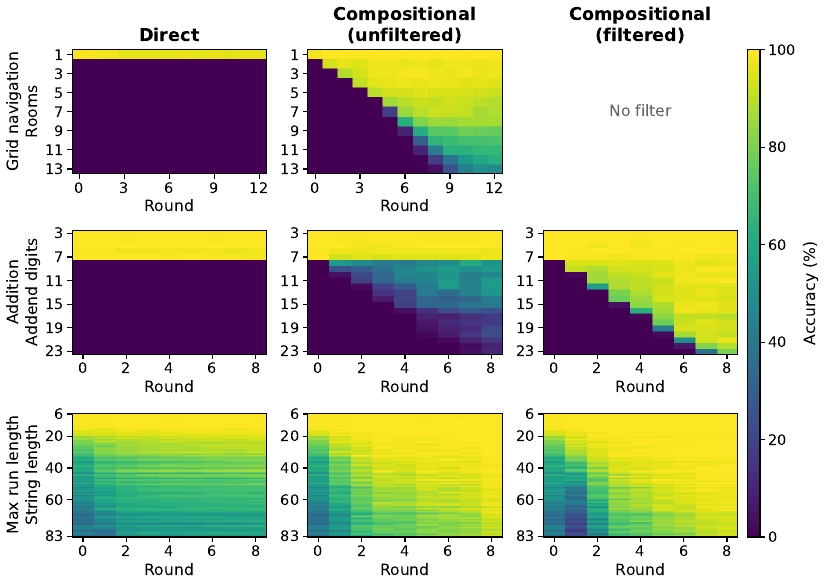}
\caption{Accuracy by instance size across self-improvement rounds for the algorithmic tasks of \Cref{subsec:main_self_improvement}, by label source. Runs in a row share the initial model, parents, and schedule. Grid navigation has no filter. The compositional panel for grid navigation and the filtered panels for addition and max run length are the runs in \Cref{fig:self_improvement_panels}; the addition panels also appear in \Cref{fig:addition_filtering}.}
\label{fig:algorithmic_baselines}
\end{figure}

\paragraph{Results.}
\Cref{fig:algorithmic_baselines} shows that direct self-improvement fails to extend reliable prediction beyond the initial training range in grid navigation or addition: after the last round, its accuracy is $0\%$ at every room count above one and every addend length above seven, while compositional self-improvement reaches $80.9\%$ mean accuracy on $2$--$13$ rooms. On max run length, the initial model already reaches $59.8\%$ mean accuracy on lengths $11$--$83$; direct self-improvement raises this to $75.9\%$, and unfiltered and filtered composition reach $97.0\%$ and $97.6\%$. \textbf{The filter has the largest effect on addition}, where unfiltered composition reaches $33.5\%$ mean accuracy on $8$--$23$-digit addends and $11.0\%$ at $23$ digits, compared with $93.7\%$ and $80.0\%$ with the filter.

\paragraph{Bash and RL warm start.}
Neither experiment filters labels by correctness. The structural check (\Cref{tab:realworld_composition_examples,tab:rl_composition_examples}) rejected none of the $42{,}556$ composed Bash labels and none of the $512$ warmup labels for strings of depth five or typed Python, so the runs in \Cref{fig:program_tasks_sft,fig:rl_frontier} are unfiltered composition. \Cref{fig:program_tasks_sft}(a) shows the Bash runs next to direct self-improvement, and \Cref{fig:rl_frontier} shows the RL runs next to direct RL. After three rounds, compositional self-improvement reaches $83$--$94\%$ accuracy on Bash pipelines of two to six commands, while direct self-improvement stays at or below $11\%$. On strings of depth five, RL after the compositional warmup reaches $19.2\%$ accuracy at $40$M tokens, and direct RL remains at $0\%$ through $80$M tokens.

\FloatBarrier

\subsection{Length-Filtered Direct Self-Improvement on Addition}
\label{appsubsec:length_filtered_direct}
\citet{lee_2025_self_improving} show that direct self-improvement can extend addition to long operands, whereas our direct baseline fails beyond the initial training range (\Cref{appsubsec:algorithmic_baselines}). Their method adds one digit per round and discards labels more than two tokens shorter than the longest prediction for the same length; they also find that forward addition, as in our setting, requires such filtering while reversed addition does not. We test whether this curriculum and length filter let direct self-improvement match compositional self-improvement.

\paragraph{Setup.}
All runs use the training setup of \Cref{tab:training_setup} and start from the same two initial models, retrained on addends of $3$--$7$ digits so that they are accurate at eight digits, as the length filter requires. We compare compositional self-improvement, adding two digits per round for eight rounds, with length-filtered direct self-improvement on the same curriculum and on the one-digit curriculum of \citet{lee_2025_self_improving} for $16$ rounds. All runs end at $23$ digits.

\begin{figure}[!htb]
\centering
\includegraphics[width=\textwidth]{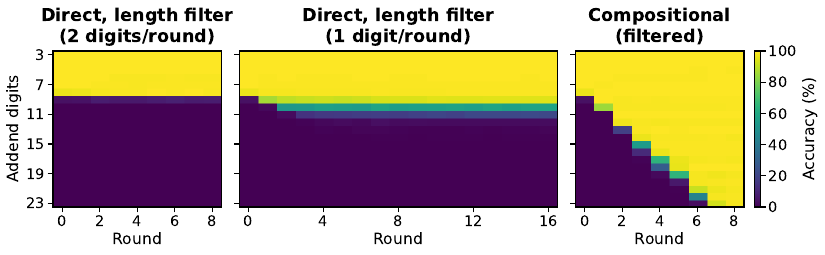}
\caption{Addition accuracy by addend length across self-improvement rounds, averaged over two retrained initial models. All runs share these initial models and end at $23$ digits; panel widths are proportional to the number of rounds. Length filter: labels shorter than the longest prediction for the same length minus two tokens are discarded \citep{lee_2025_self_improving}. Compositional runs use the boundary-carry filter of \Cref{tab:controlled_composition_examples}.}
\label{fig:length_filtered_direct}
\end{figure}

\paragraph{Results.}
\Cref{fig:length_filtered_direct} shows that \textbf{compositional self-improvement reaches $23$ digits in eight rounds, while length-filtered direct self-improvement adds at most one digit}. Compositional self-improvement reaches at least $97\%$ accuracy at every length. Direct self-improvement on the one-digit curriculum solves nine digits in the first round and no longer lengths after $16$ rounds, with about twice the GPU time.

\textbf{Direct self-improvement extends only to neighboring lengths the model already generalizes to.} The model labels nine-digit parents correctly because it generalizes from eight digits, but from ten digits on its labels are mostly wrong, and the length filter cannot detect wrong sums of the correct length. Direct self-improvement therefore depends on how far the model generalizes beyond its training range, which is small here and hard to know in advance. Compositional self-improvement instead builds labels from components the model already solves: concatenating block sums of solved lengths gives correct labels for longer addends, however far the model generalizes on its own.

\FloatBarrier

\subsection{Pipeline Length in Bash}
\label{appsubsec:bash_length}
\Cref{fig:program_tasks_sft} shows that training on pipelines of two to four commands improves accuracy on pipelines of up to six commands. This subsection examines that improvement by length and how the number of commands in the model's outputs changes across rounds.

\begin{figure}[!htb]
\centering
\includegraphics[width=\textwidth]{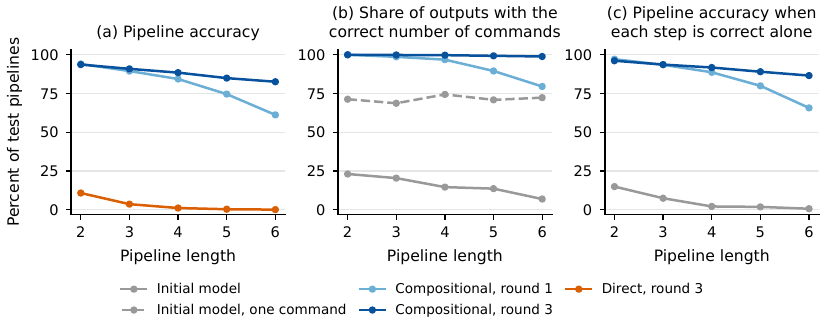}
\caption{Bash outputs on test pipelines of seen commands, by pipeline length. (a) Accuracy. (b) Share of outputs with the correct number of commands, regardless of whether the output is right, and the initial model's share of single-command outputs. (c) Accuracy on pipelines whose every step the same checkpoint writes correctly alone (see Setup). Panels (b) and (c) pool the four seeds.}
\label{fig:bash_length_analysis}
\end{figure}

\paragraph{Setup.}
We use the test pipelines of seen commands from \Cref{appsubsec:bash_setup}: $568$ two-command pipelines and $1{,}000$ pipelines at each length from three to six. For each greedy output, we count its commands and record whether the same checkpoint writes each requested command correctly when asked for it alone. Panel (c) of \Cref{fig:bash_length_analysis} uses this record. For example, the test pipeline \mbox{\texttt{head -5 | sort -r}} enters panel (c) only if the model answers \mbox{\texttt{head -5}} and \mbox{\texttt{sort -r}} when asked for each step separately; it then counts as correct only if the model's one-line answer to the full two-step instruction is also correct, which fails if it swaps the two steps or drops the \texttt{-r} flag.

\paragraph{Results.}
\Cref{fig:bash_length_analysis}(a) shows that \textbf{later rounds mainly help on the longest pipelines}. From round 1 to round 3, accuracy increases by $10.3$ points at length five and $21.4$ points at length six, compared with at most $4.1$ points at shorter lengths. Direct self-improvement stays near zero beyond two commands. \Cref{fig:bash_length_analysis}(b) explains part of the early gain: the initial model answers about $70\%$ of multi-step instructions with a single command, and one round of training on two-command pipelines teaches the model to produce the correct number of commands at most lengths. \Cref{fig:bash_length_analysis}(c) shows that at lengths four to six, accuracy also rises on pipelines for which the model writes every step correctly as a single command, so later rounds improve how the model combines these commands.

\FloatBarrier

\subsection{Transfer to Held-Out Commands in Bash}
\label{appsubsec:bash_no_correctness}
Training compositions in \Cref{subsec:extension_sft} use only $89$ of the $119$ commands; we call the other $30$ held-out commands. \Cref{fig:program_tasks_sft} shows compositional self-improvement on test pipelines with no, one, or only held-out commands. This subsection describes these test sets and adds direct self-improvement on each of them.

\paragraph{Setup.}
All $119$ commands are trained as single commands at initialization and rehearsed in every round (\Cref{appsubsec:bash_setup}). We evaluate every checkpoint on the single-command test and on pipelines of two to six commands in three test sets: no held-out command (\Cref{appsubsec:bash_length}), one held-out command ($400$ pipelines per length for lengths two and three and $1{,}000$ for four to six), and only held-out commands ($300$ and $1{,}000$ per length).

\begin{figure}[!htb]
\centering
\includegraphics[width=\linewidth]{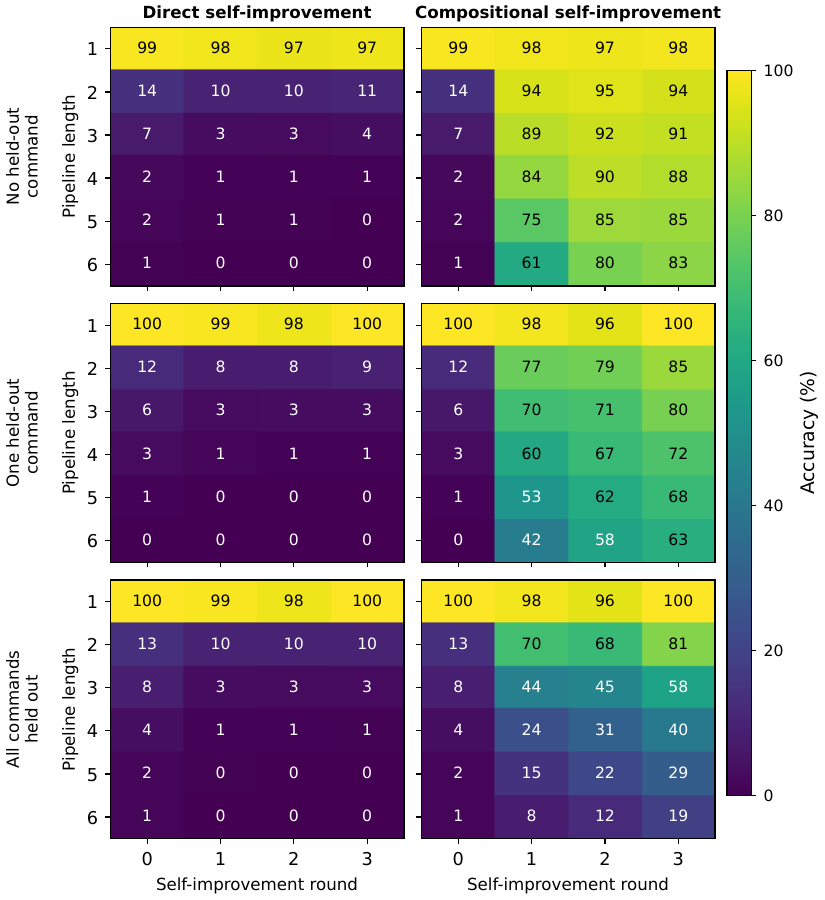}
\caption{Bash accuracy by pipeline length and self-improvement round for test pipelines with no, one, or all commands held out. The same checkpoints are evaluated on three test sets, which differ in how many of their commands were excluded from training pipelines. Length 1 is the single-command test on the 89 seen commands (top) or the 30 held-out commands (middle and bottom). Cells average four seeds; round 0 is the shared initial model.}
\label{fig:bash_transfer_sft}
\end{figure}

\paragraph{Results.}
\Cref{fig:bash_transfer_sft} shows that \textbf{composition transfers to commands that never appear in training compositions}, although less fully than to seen commands. Accuracy decreases as more commands in a pipeline are held out and as pipelines get longer, while the model still writes each held-out command correctly on its own, so the gap arises when held-out commands are composed. The third round narrows the gap, most visibly for pipelines built only from held-out commands. Direct self-improvement stays near zero on all three test sets.

\FloatBarrier

\subsection{Multi-Hop Reasoning at All Model Sizes}
\label{appsubsec:musique_sizes}
\begin{figure}[!htb]
\centering
\includegraphics[width=\textwidth]{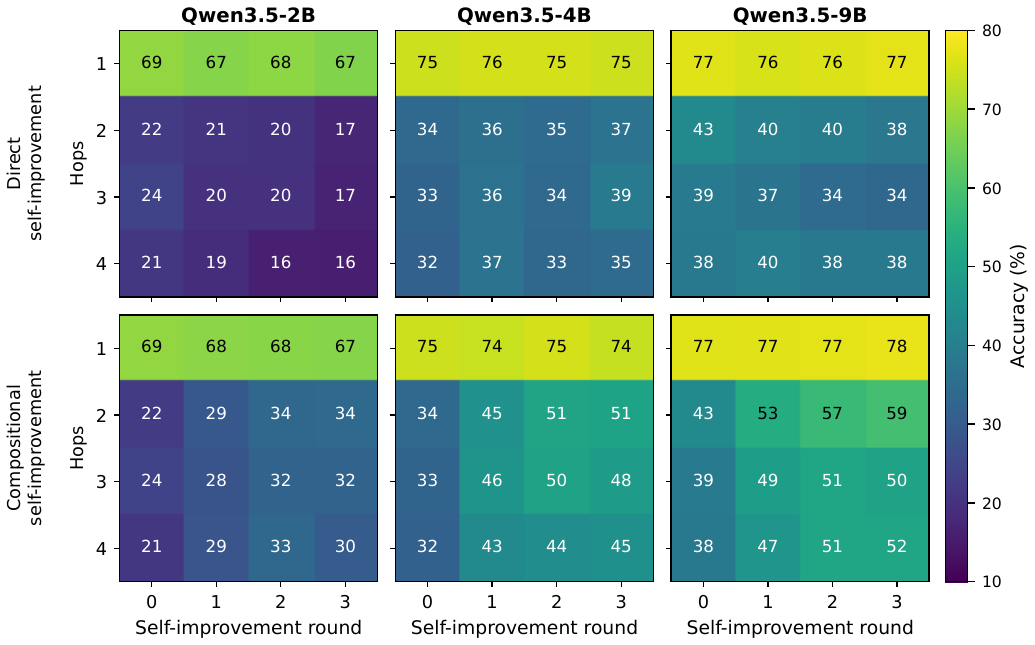}
\caption{MuSiQue results for all three model sizes. Cells show direct-answer exact-match accuracy (\%), averaged over three seeds. Top: direct self-improvement. Bottom: compositional self-improvement. The 9B panels repeat \Cref{fig:musique_hop_ladder}. After three rounds, compositional self-improvement exceeds direct self-improvement on 2- to 4-hop questions by $15.5$, $11.3$, and $17.1$ points on average for 2B, 4B, and 9B, respectively.}
\label{fig:musique_hop_ladder_sizes}
\end{figure}

\Cref{fig:musique_hop_ladder} shows per-hop accuracy for the 9B model only. \Cref{fig:musique_hop_ladder_sizes} shows it for all three model sizes.

\FloatBarrier

\subsection{Decomposition of Composed-Label Error}
\label{appsubsec:label_accuracy}\label{appsubsec:label_error_decomposition}
Compositional self-improvement helps when composed labels are more accurate than the model's direct predictions on the same parents. This subsection first compares the two label sources and then asks where the errors of composed labels come from, splitting them into the two terms of \Cref{prop:k_component_noise}: errors inherited from the components and errors introduced by the composition rule.

\paragraph{Setup.}
We audit the saved labels of each round against reference answers after training; the audit never affects which labels are kept. Following \Cref{eq:accepted_error_definitions}, all rates are computed on the accepted labels of one round:
\begin{itemize}[leftmargin=*,itemsep=1pt,topsep=2pt]
\item \emph{wrong labels} ($\rho$): the share of labels that differ from the reference answer;
\item \emph{summed component errors} ($\hat\Sigma=\sum_i\epsilon_i$): the share of labels whose $i$-th component answer is wrong ($\epsilon_i$), summed over components;
\item \emph{rule errors} ($\xi$): the share of parents whose label is wrong even with correct component answers.
\end{itemize}
In words, \Cref{prop:k_component_noise} says that wrong labels cannot exceed summed component errors plus rule errors; this holds in every measured round. A component answer is the child route in grid navigation, recovered by splitting the label at the corridor actions (on six rounds of the seed-accuracy study, where child routes were also stored, the recovery matches them); each command in Bash, run on the reference output of the preceding commands (in rounds 2 and 3, the first component is the predicted multi-command prefix); each function answer in strings and typed Python; and each step answer in MuSiQue, scored against the released step answer without aliases. For addition and max run length, component answers were not logged, so only $\rho$ and $\xi$ are available: $\xi$ is recorded during label generation for addition and measured on three regenerated candidate pools for max run length.

\paragraph{Findings.}
\begin{itemize}[leftmargin=*,itemsep=2pt,topsep=2pt]
\item \textbf{Composed labels are far more accurate than direct labels.} A composed label only requires the model to solve components it already handles, while a direct label requires it to solve the harder parent. In Bash, $91\%$ of composed labels are correct in the first round, compared with $8\%$ of direct labels, and in MuSiQue at every model size and round (\Cref{tab:musique_hop_ladder_labels}).
\item \textbf{With a valid rule, every wrong label traces back to a wrong component.} No label is wrong when all of its components are correct, and almost every wrong component produces a wrong label (\Cref{tab:label_error_decomposition}). Label quality is therefore set by the least reliable component, such as the multi-command prefix in Bash or the last functions of a string chain.
\item \textbf{In MuSiQue, errors arise mostly at the final step.} There the model answers the original question from its own earlier answers, and this step has the highest error at every size and round (\Cref{tab:musique_hop_ladder_labels}).
\item \textbf{Without a filter, addition labels are wrong almost exactly when the rule is.} The share of wrong labels matches the share of parents with a boundary carry, so nearly all label errors come from the rule rather than the components. The filter removes these parents, after which few labels are wrong.
\item \textbf{The filter keeps max run length labels within the coverage of the seed set.} A filtered label never contains a run longer than those in the seed set, as \Cref{eq:csi_output_closure} requires, so inputs with longer runs never receive composed supervision; \Cref{fig:run_length_boundary_support} evaluates such inputs.
\item \textbf{In the grid study, label quality follows the accuracy of the initial model.} Every wrong label has a wrong child route, so weaker initial models produce mostly wrong labels (\Cref{tab:grid_label_error}).
\end{itemize}

\begin{table}[!htb]
\centering\small\setlength{\tabcolsep}{3pt}
\caption{Where the errors of composed labels come from (\%), measured on the kept labels of one round. Wrong labels: share of labels that are wrong. Summed component errors: share of labels with a wrong answer in each component, summed over components. Rule errors: share of parents whose label would be wrong even if every component answer were correct. The last two columns count labels that are wrong although every component is correct, and labels that are correct although some component is wrong. Symbols refer to \Cref{prop:k_component_noise}. Addition and max run length logged no component answers, so only wrong labels and rule errors are available. Grid navigation rows cover all labels of rounds 1 and 12 (2 and 2--13 rooms), while \Cref{fig:label_accuracy_rounds} reports only the newest room count.}
\label{tab:label_error_decomposition}
\begin{tabular}{llcccccc}
\toprule
Task & Labels & Kept & \shortstack{Wrong\\labels\\{\scriptsize($\rho$)}} & \shortstack{Summed\\component\\errors {\scriptsize($\hat\Sigma$)}} & \shortstack{Rule\\errors\\{\scriptsize($\xi$)}} & \shortstack{Wrong with\\correct\\components} & \shortstack{Correct with\\a wrong\\component} \\
\midrule
Grid navigation & 2 rooms & 100 & 0.9 & 0.9 & 0.0 & 0.0 & 0.0 \\
                & 2--13 rooms & 100 & 3.5 & 3.5 & 0.0 & 0.0 & 0.0 \\
Bash & length 2 & 100 & 8.8 & 9.1 & 0.0 & 0.0 & 0.1 \\
     & length 4 & 100 & 30.3 & 32.3 & 0.0 & 0.0 & 0.3 \\
Strings & depth 5 & 100 & 23.0 & 26.8 & 0.0 & 0.0 & 0.0 \\
Typed Python & depth 3 & 100 & 5.9 & 7.4 & 0.0 & 0.0 & 1.6 \\
\midrule
Addition, unfiltered & up to 9 digits & 100 & 60.9 & -- & 60.8 & -- & -- \\
                     & up to 23 digits & 100 & 83.5 & -- & 83.3 & -- & -- \\
Addition, filtered & all rounds & 100 & 0.1--2.4 & -- & 0.0 & -- & -- \\
Max run length, unfiltered & length $\le20$ & 100 & 5.9 & -- & 5.9 & -- & -- \\
                       & length $\le83$ & 100 & 6.0 & -- & 3.1 & -- & -- \\
Max run length, filtered & all rounds & 90.0 & 0.2--1.2 & -- & 0.0 & -- & -- \\
\bottomrule
\end{tabular}
\end{table}

\begin{table}[!htb]
\centering\small\setlength{\tabcolsep}{4pt}
\caption{Training labels in the MuSiQue experiment (\%), averaged over three seeds. Label accuracy is scored with the official answer aliases. Step error gives, for each step of the composed labels in order, the share whose answer differs from the released step answer (scored without aliases); the last value is the final step. The last two rows give the share of wrong composed labels whose first wrong step is the final step, and the share of composed labels that are correct although some step answer differs from the released one. Label acceptance never uses reference answers.}
\label{tab:musique_hop_ladder_labels}
\begin{tabular}{llccc}
\toprule
Model & Quantity & Round 1 (2 hops) & Round 2 (3 hops) & Round 3 (4 hops) \\
\midrule
2B & Compositional labels correct & 44.8 & 42.3 & 49.9 \\
   & Direct labels correct & 28.9 & 21.6 & 24.3 \\
   & Step error, step 1 to final & 12.9, 58.3 & 15.9, 27.2, 61.8 & 19.7, 32.6, 33.8, 55.9 \\
   & Wrong labels, first error at final step & 83.4 & 60.7 & 39.4 \\
   & Correct labels with a wrong step & 6.8 & 15.9 & 30.8 \\
\midrule
4B & Compositional labels correct & 60.5 & 63.1 & 65.6 \\
   & Direct labels correct & 44.5 & 42.0 & 44.6 \\
   & Step error, step 1 to final & 11.1, 43.2 & 12.7, 21.6, 42.5 & 13.3, 27.3, 29.0, 43.8 \\
   & Wrong labels, first error at final step & 84.9 & 62.5 & 47.7 \\
   & Correct labels with a wrong step & 8.2 & 19.6 & 36.3 \\
\midrule
9B & Compositional labels correct & 65.2 & 66.5 & 70.5 \\
   & Direct labels correct & 53.2 & 44.7 & 46.1 \\
   & Step error, step 1 to final & 8.9, 39.3 & 11.5, 19.5, 40.0 & 13.2, 24.9, 28.2, 39.2 \\
   & Wrong labels, first error at final step & 88.1 & 63.4 & 47.0 \\
   & Correct labels with a wrong step & 8.9 & 18.6 & 36.8 \\
\bottomrule
\end{tabular}
\end{table}

\begin{table}[!htb]
\centering\small\setlength{\tabcolsep}{5pt}
\caption{Composed-label error in the grid seed-accuracy study (\Cref{fig:grid_seed_accuracy}; \%), averaged over three seeds. One-room accuracy is the initial model's test accuracy. Mean child error is $\hat\Sigma/2$ on the 2-room labels of the first round. In every cell, all wrong labels have at least one wrong child route and the rule error is $0$.}
\label{tab:grid_label_error}
\begin{tabular}{rcccc}
\toprule
Initial-training updates & One-room accuracy & Mean child error & $\rho$, 2 rooms & $\rho$, 13 rooms \\
\midrule
400 & 28.4 & 76.3 & 94.3 & 99.4 \\
1{,}500 & 58.3 & 44.1 & 68.3 & 93.7 \\
2{,}500 & 73.7 & 29.1 & 49.7 & 86.0 \\
3{,}000 & 79.9 & 21.4 & 38.1 & 77.1 \\
6{,}000 & 90.4 & 9.2 & 17.5 & 48.5 \\
8{,}000 & 95.0 & 5.3 & 10.3 & 32.2 \\
30{,}000 & 99.4 & 0.6 & 1.2 & 4.7 \\
\bottomrule
\end{tabular}
\end{table}

\FloatBarrier

\subsection{Decomposed Inference with the Initial Model}
\label{appsubsec:decomposed_initial}
Compositional self-improvement uses the decomposition only to build training labels; the trained model answers full problems in one pass. A natural alternative is to skip training and supply the decomposition at test time: the initial model answers each subproblem, and the composition rule combines its answers. This subsection compares the two.

\paragraph{Setup.}
Each test instance is split into simpler subproblems, which the initial model answers greedily; the rule then combines the answers (\Cref{appsubsec:comp_rule}). Grid chains are split into single rooms, Bash pipelines into single commands, string chains and Python problems into single functions, and MuSiQue questions into their released sub-questions. Addends are split into blocks of $3$ to $7$ digits in which no block starts with $0$, and max run length strings into $\lceil L/10\rceil$ blocks of near-equal length. The rule is applied to every test instance, including those on which it is invalid: $23.3\%$ of $8$--$23$-digit test sums have no carry across a block boundary, and $66.2\%$ of $11$--$83$-symbol test strings have no run crossing a block boundary. Test sets and metrics are those of the main experiments, and so are the initial models, except that the initial model for max run length is retrained with its recorded settings; answering directly, it reaches $98$--$100\%$ on the seed lengths (recorded: $99$--$100\%$) and $54.2\%$ on lengths $11$--$83$ (recorded: $59.8\%$). The ``Initial, direct'' columns report the recorded models.

\paragraph{Findings.}
\begin{itemize}[leftmargin=*,itemsep=2pt,topsep=2pt]
\item \textbf{Where the rule holds on every instance, decomposed inference is more accurate than the final compositional model.} The initial model already answers most components correctly, for example $118$ of the $119$ single commands, so composing its answers at test time works well (\Cref{tab:decomposed_initial,tab:decomposed_initial_rl}). The exception is 4-hop MuSiQue with the 4B and 9B models, where four sub-answers must all be correct.
\item \textbf{Where the rule fails on part of the test set, the final compositional model is more accurate.} Decomposed inference inherits every failure of the rule: in addition, it answers correctly almost exactly the sums without a carry across a block boundary, and in max run length, it errs mainly on strings with a run that continues across a block boundary.
\item \textbf{The two approaches need different information at test time.} Decomposed inference requires the reference decomposition and one model call per subproblem. Compositional self-improvement uses the decomposition only to build training labels, answers in one pass, and, with a filter, learns to handle instances on which the rule fails.
\end{itemize}

\begin{table}[!htb]
\centering\small\setlength{\tabcolsep}{4pt}
\caption{Accuracy (\%) of the initial model with direct inference and with decomposed inference, and of the final compositional model with direct inference. Decomposed inference receives the decomposition of each test instance, which direct inference does not. Rows with a size range average the accuracy per size. Bash values average four seeds for the final model; the initial model is shared. MuSiQue values average three seeds and use the 2,417 development questions of \Cref{fig:musique_hop_ladder}.}
\label{tab:decomposed_initial}
\begin{tabular}{llccc}
\toprule
Task & Test instances & Initial, direct & Initial, decomposed & Final round, direct \\
\midrule
Grid navigation & 2--13 rooms & 0.0 & 88.6 & 80.9 \\
 & 13 rooms & 0.0 & 76.5 & 47.0 \\
Addition & 8--23 digits & 0.0 & 23.2 & 93.7 \\
 & 23 digits & 0.0 & 4.0 & 80.0 \\
Max run length & 11--83 symbols & 59.8 & 87.9 & 97.6 \\
 & 83 symbols & 37.0 & 81.0 & 94.0 \\
\midrule
Bash & 2--6 commands & 5.2 & 94.9 & 88.1 \\
 & 6 commands & 0.6 & 92.9 & 82.6 \\
\midrule
MuSiQue 2B & 2 hops & 22.4 & 50.5 & 33.8 \\
 & 3 hops & 24.0 & 48.9 & 32.4 \\
 & 4 hops & 21.4 & 36.8 & 30.4 \\
MuSiQue 4B & 2 hops & 34.0 & 60.5 & 50.5 \\
 & 3 hops & 32.7 & 58.6 & 48.3 \\
 & 4 hops & 31.8 & 44.3 & 45.1 \\
MuSiQue 9B & 2 hops & 43.1 & 63.4 & 59.2 \\
 & 3 hops & 39.2 & 55.7 & 50.3 \\
 & 4 hops & 38.3 & 43.9 & 51.5 \\
\bottomrule
\end{tabular}

\end{table}

\begin{table}[!htb]
\centering\small\setlength{\tabcolsep}{4pt}
\caption{Test accuracy (\%, $512$ parents) in the RL warm-start experiments of \Cref{sec:rl_warm_start} and \Cref{appsubsec:rl_curves}. The initial model is the shared initial model; the warmup and RL columns average three seeds, after the warmup and at the last checkpoint ($40$M tokens for strings, $524$K for Python).}
\label{tab:decomposed_initial_rl}
\begin{tabular}{lcccc}
\toprule
Task & Initial, direct & Initial, decomposed & After warmup & After warmup and RL \\
\midrule
Strings, depth 5 & 0.0 & 77.3 & 11.5 & 19.2 \\
Strings, depth 4 & 0.2 & 85.4 & 38.7 & 50.1 \\
Length-preserving, depth 4 & 0.0 & 90.0 & 47.5 & 61.7 \\
Typed Python & 3.3 & 92.2 & 22.1 & 55.6 \\
\bottomrule
\end{tabular}

\end{table}

\FloatBarrier

\subsection{Pseudo-Label Accuracy by Round}
\label{appsubsec:label_accuracy_rounds}
This subsection reports the accuracy of the training labels in every round of the main experiments, audited against reference answers after training.

\paragraph{Setup.}
For the algorithmic tasks, a label is audited if its parent has the size added in that round; replay examples at the initial difficulty levels are excluded. Grid routes are scored by execution, and addition and max run length labels by exact match. Bash labels are scored on all $24$ inputs of their parent, and the RL warmup labels by the final answer (strings) or all $24$ test inputs (typed Python). MuSiQue label accuracy by round is in \Cref{tab:musique_hop_ladder_labels}.

\begin{figure}[!htb]
\centering
\includegraphics[width=\textwidth]{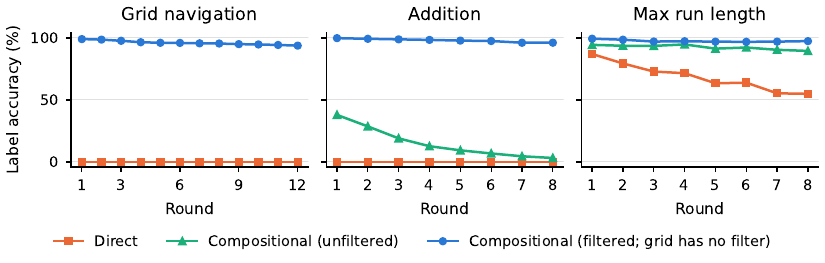}
\caption{Accuracy of the training labels for the instance size added in each round, audited against reference answers after training. Grid navigation has no filter.}
\label{fig:label_accuracy_rounds}
\end{figure}

\paragraph{Results.}
\textbf{Composed labels remain more accurate than direct labels in every round.} In grid navigation, $99.1\%$ of composed labels are correct in round 1 and $93.8\%$ in round 12, while every direct label is wrong (\Cref{fig:label_accuracy_rounds}). With the filter, max run length and addition labels stay above $96\%$ in every round. Without it, max run length label accuracy decreases from $94.5\%$ to $89.5\%$ and addition label accuracy from $38.0\%$ to $3.2\%$, since the share of boundary carries grows with the number of blocks. Direct max run length labels decrease from $87.1\%$ to $54.9\%$, and direct addition labels are wrong in every round. In Bash (\Cref{tab:label_accuracy_rounds}), composed label accuracy decreases from $90.6\%$ in round 1 to $80.6\%$ in round 3, and direct label accuracy from $7.9\%$ to $0.5\%$. The RL warmup labels are correct for $77.0\%$ of parents for strings of depth five and $94.1\%$ for typed Python.

\begin{table}[!htb]
\centering\small\setlength{\tabcolsep}{5pt}
\caption{Accuracy (\%) of training labels. (a) Bash, by self-improvement round, averaged over four seeds. (b) The $512$ labels of the single compositional warmup in each RL configuration.}
\label{tab:label_accuracy_rounds}
\begin{minipage}[t]{0.47\linewidth}
\centering
(a) Bash\par\smallskip
\begin{tabular}{lccc}
\toprule
Label source & Round 1 & Round 2 & Round 3 \\
\midrule
Compositional & 90.6 & 83.6 & 80.6 \\
Direct & 7.9 & 2.0 & 0.5 \\
\bottomrule
\end{tabular}

\end{minipage}\hfill
\begin{minipage}[t]{0.47\linewidth}
\centering
(b) RL warmup\par\smallskip
\begin{tabular}{lc}
\toprule
Configuration & Warmup labels \\
\midrule
Strings, depth 5 & 77.0 \\
Strings, depth 4 & 87.5 \\
Length-preserving, depth 4 & 87.7 \\
Typed Python & 94.1 \\
\bottomrule
\end{tabular}

\end{minipage}
\end{table}

\FloatBarrier

\subsection{Additional RL Warm-Start Experiments}
\label{appsubsec:rl_curves}
\Cref{fig:rl_frontier} reports results for depth-five string transformations and typed Python programs. Here, we repeat the RL comparison on two depth-four string configurations: chains
drawn from all $21$ functions and chains restricted to the $11$ length-preserving functions.

\paragraph{Setup.}
Length-preserving functions leave the input string length unchanged, including rotations, character shifts, and case changes (\Cref{appsubsec:comp_rule}). Both configurations follow the training and evaluation protocol of \Cref{appsubsec:rl_setup}. We evaluate CSI+RL after the compositional warmup and at $40$M tokens, and compare it with direct RL trained for up to $80$M tokens.

\begin{figure}[!htb]
\centering
\includegraphics[width=0.8\textwidth]{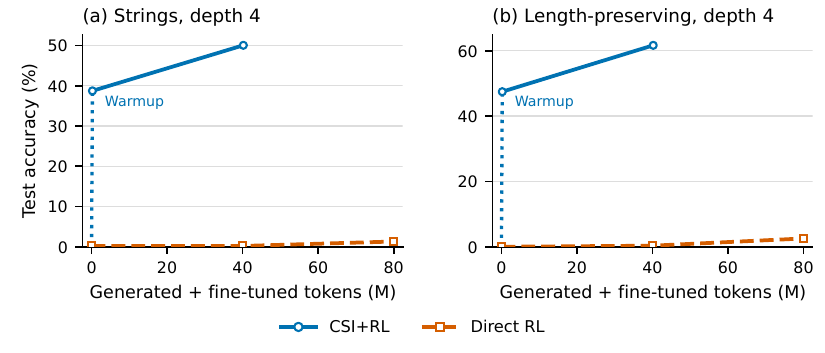}
\caption{Test accuracy on 512 held-out problems for two depth-four string configurations: (a) all $21$ functions and (b) the $11$ length-preserving functions. Results are averaged over three seeds. The dotted segment denotes the compositional SFT warmup; token accounting follows
\Cref{fig:rl_frontier}.}
\label{fig:rl_frontier_appendix}
\end{figure}

\paragraph{Results.}
\Cref{fig:rl_frontier_appendix} shows a similar pattern to depth-five strings. The compositional warmup already achieves approximately $40\%$ accuracy with all functions and $48\%$ with length-preserving functions. Subsequent RL improves accuracy to approximately $50\%$ and $61\%$, respectively. In contrast, direct RL remains below $3\%$ accuracy even with twice the token budget. These results further show that compositional supervision can provide a useful starting point for RL when correct rollouts are otherwise rare.

\FloatBarrier

\end{document}